\documentclass[preprint,5p,authoryear]{elsarticle}

\usepackage{amsmath,amssymb,amsfonts}
\usepackage{algpseudocode,algorithm}
\usepackage{subfigure}
\usepackage{graphicx}

\journal{Neural Networks}

\begin{document}

\sloppy

\begin{frontmatter}



\title{t-Soft Update of Target Network for Deep Reinforcement Learning}


\author{Taisuke Kobayashi\corref{cor}}
\ead{kobayashi@is.naist.jp}
\ead[url]{http://kbys\_t.gitlab.io/en/}

\author{Wendyam Eric Lionel Ilboudo\corref{}}
\ead{ilboudo.wendyam\_eric.in1@is.naist.jp}

\cortext[cor]{Corresponding author}

\address{Nara Institute of Science and Technology, Nara, Japan}

\begin{abstract}

This paper proposes a new robust update rule of target network for deep reinforcement learning (DRL), to replace the conventional update rule, given as an exponential moving average.
The target network is for smoothly generating the reference signals for a main network in DRL, thereby reducing learning variance.
The problem with its conventional update rule is the fact that all the parameters are smoothly copied with the same speed from the main network, even when some of them are trying to update toward the wrong directions.
This behavior increases the risk of generating the wrong reference signals.
Although slowing down the overall update speed is a naive way to mitigate wrong updates, it would decrease learning speed.
To robustly update the parameters while keeping learning speed, a t-soft update method, which is inspired by student-t distribution, is derived with reference to the analogy between the exponential moving average and the normal distribution.
Through the analysis of the derived t-soft update, we show that it takes over the properties of the student-t distribution.
Specifically, with a heavy-tailed property of the student-t distribution, the t-soft update automatically excludes extreme updates that differ from past experiences.
In addition, when the updates are similar to the past experiences, it can mitigate the learning delay by increasing the amount of updates.
In PyBullet robotics simulations for DRL, an online actor-critic algorithm with the t-soft update outperformed the conventional methods in terms of the obtained return and/or its variance.
From the training process by the t-soft update, we found that the t-soft update is globally consistent with the standard soft update, and the update rates are locally adjusted for acceleration or suppression.

\end{abstract}




\end{frontmatter}


\section{Introduction}

In the last decade, since deep neural networks (DNNs) have recorded a better-than-human image recognition accuracy~\citep{krizhevsky2012imagenet}, expectations for their high function approximation capabilities have increased.
Hidden relationship between data, which used to be a black box, has been revealed by solving estimation problems with DNNs.
Indeed, many applications have been reported in various fields such as economics~\citep{pang2020innovative} and engineering~\citep{naderpour2020bio,panyafong2020heat}.

As a new frontier for DNNs after the estimation problems, control problems with black-box relationship between observed states and control commands are in the spotlight recently.
Reinforcement learning (RL)~\citep{sutton2018reinforcement} and its extension using DNNs to approximate policy and value functions, named deep reinforcement learning (DRL)~\citep{silver2016mastering,levine2018learning}, is one of the promising methodologies for solving the control problems.
For example, human-robot interaction~\citep{modares2015optimized} and deformable object manipulation~\citep{tsurumine2019deep} have been tackled.

However, in practice, only applying DNNs as function approximators would make learning process unstable due to their high nonlinearity.
To stably learn the optimal policy, therefore, techniques to reduce the variance of learning signals have been developed for DRL.
For example, the experience replay~\citep{lin1992self,andrychowicz2017hindsight} method can statistically mitigate the effects of anomalies by allowing DRL to transform into mini-batch learning;
and various regularization techniques for the policy~\citep{schulman2017proximal,haarnoja2018soft,kobayashi2019student,parisi2019td} yield a conservative learning algorithm that avoids updates into the wrong directions;
also, by learning two ensemble value functions, either one can be selected to minimize the approximation bias~\citep{fujimoto2018addressing}.
Alternatively, several heuristic ways, such as reward and/or gradient clipping, have also been employed in many cases.

As one of such techniques employed for stable learning, target networks have been proposed~\citep{mnih2015human}.
The target network generates the reference signals for the paired main network, and ``slowly'' updates its parameters (i.e., weights and biases in the network) toward the parameters of the main network.
In this way, the reference signals would not fluctuate frequently, thereby making it easier to learn more stably, although the learning speed of models employing the target network is basically decreased~\citep{kim2019deepmellow}.
When the target network was first introduced, the ``hard'' update strategy, which copies, every few steps, the main network into the target one, was used.
However, after that, a new strategy, the ``soft'' update, by which the new parameters for the target network are interpolated, through a fixed ratio, between the current parameters of the target network and the parameters of the main network, became the mainstream in DRL libraries probably due to its flexibility~\citep{stooke2019rlpyt,fujita2019chainerrl}.

As another problem different from the slowdown of learning speed, the soft update is basically sensitive to noise and outliers in the updates of the parameters for the main network.
For example, even if parts of the parameters for the main network are updated largely along some wrong directions, the soft update will approve all of them without any checks, and copy them into the target network using the fixed ratio.
A naive solution for this problem would be to make the ratio for copy as small as possible, but it will slow down the learning speed further.
This means that a trade-off between the sensitivity to noise and outliers and learning speed is given by the fixed ratio employed for copying.

As pointed out in the literature~\citep{ilboudo2020tadam}, this problem comes from the exponential moving average (EMA), which can be regarded as the update rule of the mean of a normal distribution with a fixed number of samples.
Therefore, based on that literature, this paper proposes a new update rule of the target network, named ``t-soft'' update.
It is inspired by student-t distribution, which is a well-known distribution robust to outliers~\citep{tipping2005variational,shah2014student,kobayashi2019student}.
The EMA in the soft update is replaced with a moving average derived from the location parameter of the student-t distribution.
In addition, its computation and memory costs are minimized as much as possible by assuming a simple stochastic model while keeping the performance of the t-soft update.
Note that the properties of the student-t distribution are still inherited even with such a approximated implementation.

We verify the superiority of the t-soft update through four kinds of dynamical simulations for DRL.
Here, PyBullet robotics simulations~\cite{coumans2016pybullet,brockman2016openai} are employed as benchmarks, and an online actor-critic algorithm with DNNs is combined with the proposed and conventional update rules of target network.
The simulation results verify that the t-soft update outperforms the conventional methods in terms of the obtained return and/or its variance.
In addition, the training process by the t-soft update is analyzed to confirm that the t-soft update is globally consistent with the standard soft update, and the update rates are locally adjusted for acceleration or suppression.

The remainder of this paper is organized as follows.
Section~\ref{sec:preliminary} introduces the basics for DRL with the target network.
Section~\ref{sec:proposal} proposes the t-soft update and its practical implementation so that it inherits the properties of the student-t distribution.
Section~\ref{sec:simulation} evaluates the improvement of learning performance by the t-soft update compared to the conventional methods.
Section~\ref{sec:conclusion} concludes this paper with a summary and future work.

\section{Preliminaries}
\label{sec:preliminary}

\subsection{Reinforcement learning}

RL enables an agent to learn the optimal policy, which can achieve the maximum sum of rewards from an environment~\citep{sutton2018reinforcement}.
In RL, Markov decision process (MDP) with the tuple $(\mathcal{S}, \mathcal{A}, \mathcal{R}, p_1, p_T, \gamma)$ is assumed.

After getting the initial state $s_1 \in \mathcal{S}, s_1 \sim p_1(s_1)$, the agent decides the action at the time step $t \in \mathbb{N}$, $a_t \in \mathcal{A}$, using the policy $a_t \sim \pi(a_t \mid s_t)$.
By performing the action $a_t$ on the environment, the state is transited to the next according to the transition probability, $s_{t+1} \sim p_T(s_{t+1} \mid s_t, a_t)$.
At the same time, the agent gets a reward according to the reward function: $r_t = r(s_t, a_t, s_{t+1}) \in \mathcal{R}$.

The sum of rewards is converted to produce the return defined as $R_t = \sum_{k=0}^\infty \gamma^k r_{t+k}$ with a discount factor $\gamma \in [0, 1)$ for finiteness.
As already mentioned, the purpose of the agent is to acquire the optimal policy $\pi^*$ that maximizes $R_t$.
To this end, numerous algorithms have been proposed, and in this paper, we basically employ an actor-critic algorithm~\citep{williams1992simple,peters2008natural}, which is suitable for robotics due to the capability of directly optimizing the policy with a parameters set $\eta$ in continuous action space (see details in~\ref{app:actor_critic}).

\subsection{Target network with soft update}

\begin{figure}[tb]
    \centering
    \includegraphics[keepaspectratio=true,width=0.98\linewidth]{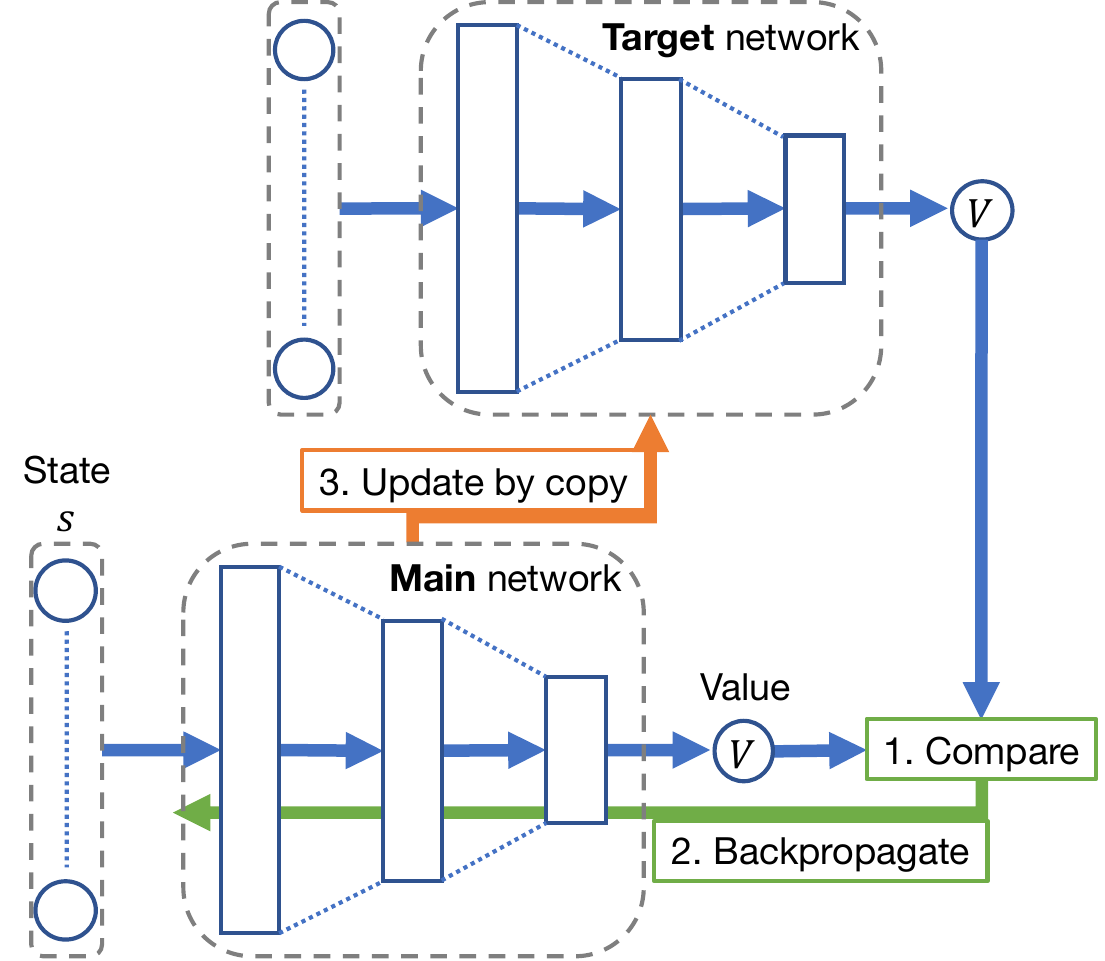}
    \caption{Target network for stable learning of value function:
        the target network gives the reference signals for the main network;
        the loss between the main and target networks is backpropagated into the main network for optimization;
        the main network is softly copied into the target network.
    }
    \label{fig:target_net}
\end{figure}

In most of DRL methods using value function (i.e., the expectation of the return), $Q(s,a)$ and/or $V(s)$, the target network is employed to reduce the variance of learning signals~\citep{mnih2015human,stooke2019rlpyt,fujita2019chainerrl} (see Fig.~\ref{fig:target_net}).
For example, when the advantage function $A(s,a) = Q(s,a) - V(s) = r + \gamma V(s^\prime) - V(s)$ is introduced as the learning signal, the following loss is minimized by optimizing the parameters set for the main network $\theta$:
\begin{align}
    \mathcal{L}(\theta) = (r + \gamma V(s^\prime; \phi) - V(s; \theta))^2
    \label{eq:loss_value}
\end{align}
where $\phi$ denotes the parameters set for the target network.
That is, the target network outputs the supervision for the bootstrap learning.
Another target network $\zeta$ as a baseline policy can be prepared for sampling actions instead of the policy $\eta$ and for smoothly optimizing $\eta$ (see~\ref{app:actor_critic}).
Note that this paper describes only the pair of $\theta$ and $\phi$ without loss of generality.

The above minimization problem for $\theta$ is solved basically according to first-order gradients with some of stochastic gradient descent (SGD) optimizers~\citep{ziyin2020laprop,ilboudo2020tadam,kobayashi2020towards}.
However, it does not update $\phi$ directly, and therefore, an alternative update rule for $\phi$ is needed.
While its optimal value is expected to be $\theta$ after its update, it should be noted that the update by SGDs is unstable in the optimization of nonlinear function approximations like deep learning.

Hence, instead of immediately following $\theta$, we use the soft update rule, which smoothly updates $\phi$ to $\theta$:
\begin{align}
    \phi \gets (1 - \tau) \phi + \tau \theta
    \label{eq:soft_update}
\end{align}
where $\tau$ denotes the smoothness, and if $\tau = 1$, this update rule becomes the hard update.

\section{Proposal: t-soft update}
\label{sec:proposal}

\subsection{Analogy between EMA and normal distribution}

The soft update employs the EMA of $\theta$, i.e., $\phi$ is regarded as its moving average.
As explained in the literature~\citep{ilboudo2020tadam}, the EMA extracts the same ratio from the new value $\theta$ even if it contains noise and outliers (i.e., sudden changes by extreme gradients).
To avoid the adverse effects of noise and outliers during the update, we should understand the EMA more deeply.

To this end, let us focus on the analogy between the EMA and normal distribution as well as the above literature did.
Specifically, given the $N$ sampled data $\{x_n\}_{n=1}^N$, the maximum likelihood estimation of normal distribution derives its mean $\mu$ as follows:
\begin{align}
    \mu_N &= \cfrac{1}{N} \sum_{n=1}^N x_n
    \nonumber \\
    &= \cfrac{1}{N} \sum_{n=1}^{N-1} x_n + \cfrac{1}{N} x_N
    \nonumber \\
    &= \cfrac{N-1}{N} \left( \cfrac{1}{N-1} \sum_{n=1}^{N-1} x_n \right) + \cfrac{1}{N} x_N
    \nonumber \\
    &= (1 - \tau_N) \mu_{N-1} + \tau_N x_N
    \label{eq:mean_normal}
\end{align}
where $\tau_N = 1/N$.
When the effective number of sampled data is fixed ($N = \mathrm{const.} \in \mathbb{N}$ and $\tau_N = \mathrm{const.} \in (0, 1]$), eq.~\eqref{eq:mean_normal} matches the EMA (i.e., the soft update).
Since normal distribution is well known as a distribution sensitive to outliers, we agree that the sensitivity of the EMA is taken over from it.
This sensitivity can be expected from the fact that the mean $\mu$ is gained by treating all data equivalently.

\subsection{Moving average based on student-t distribution}

The concept of our proposal stands on replacing normal distribution to the distribution robust to outliers.
As such a distribution, student-t distribution is employed in this paper following the previous studies~\citep{tipping2005variational,shah2014student,kobayashi2019student}.
Hence, to derive a new update rule based on student-t distribution, we derive the alternative moving average formula from the maximum likelihood estimation of student-t distribution.

Suppose $d$-dimensional diagonal student-t distribution with model parameters, i.e., location $\mu \in \mathbb{R}^d$, scale $\sigma \in \mathbb{R}_+^d$, and degrees of freedom $\nu \in \mathbb{R}_+$.
Given the $N$ sampled data $\{x_n\}_{n=1}^N$, the estimated $\mu_N$ is derived as follows:
\begin{align}
    \mu_N = \cfrac{1}{W_N} \sum_{n=1}^N w_n x_n
\end{align}
where
\begin{align}
    W_N &= \sum_{n=1}^N w_n
    \label{eq:t_sum} \\
    w_n &= \cfrac{\nu + d}{\nu + (x_n - \mu_{N-1})^2\sigma^{-2}} = \cfrac{\nu + d}{\nu + D}
    \label{eq:t_weight}
\end{align}
where $D = (x_n - \mu_{N-1})^2 \sigma^{-2}$.
That is, since $\mu_{N-1}$ remains inside of the formula, $\mu_N$ has to be computed recursively.

This fact requires us to approximate the derivation of moving average as follows:
\begin{align}
    \mu_N &\simeq \cfrac{W_{N-1}}{W_{N-1} + w_N}\cfrac{1}{W_{N-1}} \sum_{n=1}^{N-1} w_n x_n + \cfrac{w_N}{W_{N-1} + w_N} x_N
    \nonumber \\
    &\simeq \cfrac{W_{N-1}}{W_{N-1} + w_N} \mu_{N-1} + \cfrac{w_N}{W_{N-1} + w_N} x_N
    \nonumber \\
    &= (1 - \tau_w) \mu_{N-1} + \tau_w x_N
    \label{eq:mean_student}
\end{align}
where $\tau_w = w_N / (W_{N-1} + w_N) \in (0, 1)$.
We notice that $W_N \simeq W_{N-1} + w_N$ and $W_{N-1}^{-1} \sum_{n=1}^{N-1} w_n x_n \simeq \mu_{N-1}$ are the approximated terms.
If the update of the model parameters are slow enough, these approximations are with high precision.

As can be seen in eq.~\eqref{eq:mean_student}, it becomes the same form as eq.~\eqref{eq:mean_normal} with a different definition of $\tau$.
While $\tau$ in the case of normal distribution is fixed to be constant, $\tau$ in the case of student-t distribution is adaptive even if the effective number of samples $N$ is fixed.
For example, if the new sample $x$ is far away from the current $\mu$, $w$ will be small with large $D$, and $x$ hardly affects the update of $\mu$.
Otherwise, $x$ strongly affects the update of $\mu$ according to the large $w$.
This behavior is desired for the robust and efficient update of the target network.

\subsection{Practical design of t-soft update}

\begin{algorithm}[tb]
    \caption{Proposed t-soft update with hyperparameters $\nu \in \mathbb{R}_+$ and $\tau \in (0, 1]$}
    \label{alg:proposal}
    \begin{algorithmic}[1]
        \State{Initialize $\theta$}
        \State{$\phi \gets \theta$}
        \State{Initialize optimizer SGD with learning rate $\alpha$}
        \State{$W_i = (1 - \tau) \tau^{-1}$}
        \State{$\sigma_i = \epsilon \ll 1$}
        \While{True}
            \State{Compute $\mathcal{L}(\theta)$}
            \State{$\theta \gets \theta - \alpha \mathrm{SGD}(\nabla_\theta \mathcal{L})$}
            \If{Meet update interval}
                \For{$\theta_i, \phi_i \subset \theta, \phi$}
                    \Comment{Extract $i$-th subset}
                    \State{$\Delta_i^2 = \frac{1}{N_i} \sum_{n=1}^{N_i}(\theta_{i,n} - \phi_{i,n})^2$}
                    \State{$w_i = (\nu + 1)(\nu + \Delta_i^2 \sigma_i^{-2})^{-1}$}
                    \State{$\tau_i = w_i (W_i + w_i)^{-1}$}
                    \State{$\phi_i \gets (1 - \tau_i) \phi_i + \tau_i \theta_i$}
                    \State{$\tau_{\sigma_i} = \tau w_i \nu (\nu + 1)^{-1}$}
                    \State{$\sigma_i^2 \gets (1 - \tau_{\sigma_i}) \sigma_i^2 + \tau_{\sigma_i} \Delta_i^2$}
                    \State{$W_i \gets (1 - \tau) (W_i + w_i)$}
                \EndFor
            \EndIf
        \EndWhile
    \end{algorithmic}
\end{algorithm}

\begin{figure}[tb]
    \centering
    \includegraphics[keepaspectratio=true,width=0.98\linewidth]{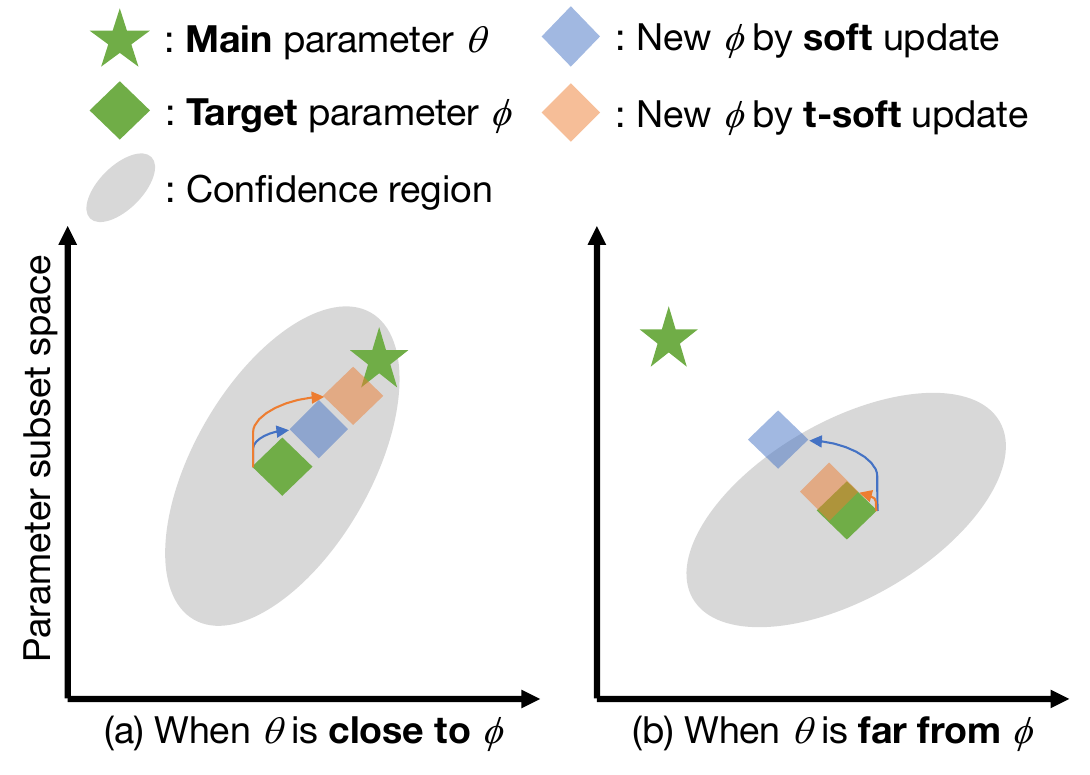}
    \caption{Qualitative behavior of the t-soft update rule:
        (a) when $\theta$ is close to $\phi$, the update rate of the t-soft update is larger than $\tau$, that accelerates the update of $\phi$ towards $\theta$;
        (b) when $\theta$ is far from $\phi$, the update of $\phi$ is suppressed since this rule suspects that $\theta$ is wrong.
    }
    \label{fig:tsoft_behavior}
\end{figure}

\begin{table*}[tb]
    \caption{Simulation environments provided by Pybullet Gym~\citep{brockman2016openai,coumans2016pybullet}}
    \label{tab:sim_environment}
    \centering
    \begin{tabular}{ccccc}
        \hline\hline
        ID & Name & State space $d_s$ & Action space $d_a$ & Episode $E$ \\
        \hline
        InvertedPendulumSwingupBulletEnv-v0 & Swingup & 5 & 1 & 150 \\
        InvertedDoublePendulumBulletEnv-v0 & DoublePendulum & 9 & 1 & 2000 \\
        HalfCheetahBulletEnv-v0 & HalfCheetah & 26 & 6 & 1500 \\
        AntBulletEnv-v0 & Ant & 28 & 8 & 1500 \\
        \hline\hline
    \end{tabular}
\end{table*}

\begin{table*}[tb]
    \caption{Common hyperparameters for the simulations}
    \label{tab:sim_parameter}
    \centering
    \begin{tabular}{ccc}
        \hline\hline
        Symbol & Meaning & Value \\
        \hline
        $N$ & Number of neurons & 100 \\
        $L$ & Number of layers & 5 \\
        $\gamma$ & Discount factor & 0.99 \\
        $\alpha$ & Learning rate & 3e-4 \\
        $(\lambda_\mathrm{max}^1, \lambda_\mathrm{max}^2, \kappa)$ & Decaying factors for adaptive eligibility traces~\citep{kobayashi2020adaptive} & (0.5, 0.95, 10) \\
        $\epsilon$ & Threshold for regularization of policy update~\citep{kobayashi2020proximal} & 0.1 \\
        $\beta_{DE}$ & Gain for maximization of policy entropy~\citep{haarnoja2018soft} & 0.01 \\
        $\beta_{TD}$ & Gain for TD regularization~\citep{parisi2019td} & 0.01 \\
        \hline\hline
    \end{tabular}
\end{table*}

Although the new moving average in eq.~\eqref{eq:mean_student} is effective to ignore outliers, to employ it as the update rule of target network, we have to design the parameters $\sigma$, $\nu$, $d$, and $W$.
In particular, unlike the literature~\citep{ilboudo2020tadam}, where $\sigma$ is already estimated via another EMA, the update rule of $\sigma$ is additionally required.
It is also desirable to minimize the computation and memory costs to the same level as the conventional soft update.
To this end, this paper proposes a practical design of the t-soft update.

First of all, we define two hyperparameters, $(\tau, \nu)$, where $\tau \in (0, 1]$ and $\nu > 0$.
Suppose that the update is performed for each subset of the parameters set (a.k.a., weights or a bias of each layer).
In that case, the $i$-th subset has $\sigma_i$, $d_i$, and $W_i$.

To reduce the memory cost to store $\sigma$ for all the parameters, each $i$-th subset is assumed to have a common $\sigma_i \in \mathbb{R}_+$.
This means that for each subset, a $d_i = d = 1$-dimensional student-t distribution is assumed.
Following this assumption, eq.~\eqref{eq:t_weight} is redefined as follows:
\begin{align}
    w_i = \cfrac{\nu + 1}{\nu + \Delta_i^2\sigma_i^{-2}}
    \label{eq:tsoft_weight}
\end{align}
where
\begin{align}
    \Delta_i^2 = \cfrac{1}{N_i}\sum_{n=1}^{N_i}(\theta_{i,n} - \phi_{i,n})^2
    \label{eq:tsoft_diff}
\end{align}
where $N_i$, $\theta_i$, and $\phi_i$ denote the number of parameters in $i$-th subset, the main and target network's parameters in $i$-th subset, respectively.
The range of $w_i$ is theoretically $(w_i |_{\Delta_i^2 \to \infty} , w_i |_{\Delta_i = 0}] = (0, (\nu + 1) / \nu]$, although $\Delta_i^2 \to \infty$ would not occur in practice due to the finite gradients for updating the main network.

For finiteness of $W_i$, the past data is assumed to be decayed as time goes on.
Specifically, when the decaying rate is given as $1 - \tau$ for simplicity, $W_i$ is given as follows:
\begin{align}
    W_{i,t} &= \sum_{k=0}^{t-1} (1 - \tau)^k w_{i,t-k}
    \nonumber \\
    W_i &\gets (1 - \tau) (W_i + w_i)
    \label{eq:tsoft_sum}
\end{align}
where the initial value of $W_i$ is $(1 - \tau) / \tau$.
This design expects that the t-soft update reverts to the soft update when $\nu \to \infty$ as the student-t distribution reverts to the normal distribution in that case.

Based on eq.~\eqref{eq:mean_student}, the t-soft update is derived as follows:
\begin{align}
    \phi_i \gets (1 - \tau_i) \phi_i + \tau_i \theta_i
    \label{eq:tsoft_upd}
\end{align}
where $\tau_i = w_i / (W_i + w_i) \in (0, 1)$.
That is, the update of $i$-th subset is suppressed if the mean of the differences between $\theta_i$ and $\phi_i$ (i.e., $\Delta_i$) is larger than the threshold implied by $\sigma_i$.
Note that, due to $\tau_i \in (0, 1]$, the target network can eventually converge to the main network even with the t-soft update.

After that, $\sigma_i$ is adjusted according to the observed $\Delta_i$.
According to the gradient for maximum likelihood (see details in~\ref{app:student_sigma}), the following update is derived.
\begin{align}
    \sigma_i^2 &\gets \left (1 - \tau w_i \cfrac{\nu}{\nu + 1} \right ) \sigma_i^2 + \tau w_i \cfrac{\nu}{\nu + 1} \Delta_i^2
    \nonumber \\
    &= (1 - \tau_{\sigma_i}) \sigma_i^2 + \tau_{\sigma_i} \Delta_i^2
    \label{eq:tsoft_var}
\end{align}
Here, $\sigma_i^2$ can be regarded as the moving average of $\Delta_i^2$ with $\tau_{\sigma_i} = \tau w_i \nu/(\nu + 1)$ update rate.
Note that, due to $w_i \in (0, (\nu + 1) / \nu]$, the update rate is smaller than $\tau$ and converges to $\tau$ when $\nu \to \infty$.
If a new observation deviates from the previous experience (i.e., $\Delta_i^2 > \sigma_i^2$), this update rule suppresses the update of $\sigma_i^2$ towards $\Delta_i^2$.
Hence, the effects of outliers would be mitigated in subsequent updates.

In summary, the proposed t-soft update is implemented as described in Alg.~\ref{alg:proposal}.
Note that a SGD optimizer and a loss function for RL, $\mathcal{L}$, can be arbitrarily selected.
In addition, the behavior of the t-soft update is visualized in Fig.~\ref{fig:tsoft_behavior}.
This behavior yields robust and efficient update of the target network.

\section{Simulations}
\label{sec:simulation}

\subsection{Benchmark tasks}

Four benchmark tasks for DRL simulated by Pybullet Gym~\citep{brockman2016openai,coumans2016pybullet} are prepared, and listed in Table~\ref{tab:sim_environment}.
Their rewards are designed based on the following purposes:
\begin{enumerate}
    \renewcommand{\labelenumi}{(\alph{enumi})}
    \item Swingup:
    A cart swings up a pole and keeps it standing.
    \item DoublePendulum:
    A cart keeps a double pendulum standing.
    \item HalfCheetah:
    A two-dimensional cheetah with two legs walks as fast as possible.
    \item Ant:
    A three-dimensional quadruped walks as fast as possible.
\end{enumerate}

After learning, the agent performs the learned task 50 times to compute the sum of rewards for each, and their median is used as the score.
This evaluation process is for eliminating the effects of random initial states and for evaluating the robustness of the learned policy.
In total, 10 trials are performed for each condition, and each trial sets the random seed as the trial number.

\subsection{Network architecture}

Basic network architecture, which is implemented by PyTorch~\citep{paszke2017automatic}, has $L$ fully connected intermediate layers with $N$ neurons with layer normalization~\citep{ba2016layer} and Swish activation function~\citep{ramachandran2017swish,elfwing2018sigmoid}.
Using this network architecture, the value function $\theta$ and the policy $\eta$ for the actor-critic algorithm (see details in the next section and~\ref{app:actor_critic}) are implemented.

In the simulations, $L=5$ and $N=100$ are set empirically.
These values are determined to ensure the sufficient ability for function approximation while avoiding overfitting.
The previous work~\citep{kobayashi2020proximal} indicated that these values could enable the agent to learn all the given tasks.

For optimization of the above neural networks by SGD, a robust optimizer (i.e., a combination of LaProp~\citep{ziyin2020laprop}, t-momentum~\citep{ilboudo2020tadam}, and d-AmsGrad~\citep{kobayashi2020towards}) is employed with their default parameters except the learning rate $\alpha$.
Thanks to this optimizer, the effects of noisy learning signals caused by bootstrap leaning like DRL can be reduced without numerous data.
Note that since the wrong updates cannot be removed perfectly even by this optimizer, following the main network at the constant speed (i.e., by the hard and soft updates) may deteriorate learning performance.

\subsection{Learning methods}

In the simulations, we employ the actor-critic algorithm~\citep{williams1992simple,peters2008natural} as the main framework to learn the optimal policy.
According to the latest studies, it is customized for stable and efficient learning as follows.

First, the policy is explicitly modeled as the multivariate diagonal student-t distribution~\citep{kobayashi2019student} with the location $\mu \in \mathbb{R}^{d_a}$; the scale $\sigma \in \mathbb{R}_+^{d_a}$; and the degree of freedom $\nu \in \mathbb{R}_+$.
These parameters are approximated by the above neural network with $\eta$.

To accelerate learning speed, the adaptive eligibility traces method~\citep{kobayashi2020adaptive} is implemented with the following hyperparameters: $(\lambda_\mathrm{max}^1, \lambda_\mathrm{max}^2, \kappa)$.
This method is useful for online learning without experience replay~\citep{lin1992self,andrychowicz2017hindsight}, where non-stationary robotic tasks are naturally allowable.

To stably learn the tasks (i.e., to reduce the variance of learning results), the following latest regularization techniques are also combined.
Specifically, a proximal policy optimization with relative Pearson divergence (PPO-RPE)~\citep{kobayashi2020proximal} with a threshold value $\epsilon$ is employed to regularize the divergence between the policy with $\eta$ and the baseline policy with $\zeta$.
To avoid local optima, a soft actor-critic (SAC)~\citep{haarnoja2018soft} introduces the policy entropy regularization with a regularization weight $\beta_{DE}$.
In addition, a temporal difference (TD) regularization~\citep{parisi2019td} with a regularization weight $\beta_{TD}$ is combined for stable interaction between the actor and the critic.
Note that, we found that the regularization by the absolute value of TD error is more stable than the original implementation (i.e., its square) due to task-specific range of TD error.

Table~\ref{tab:sim_parameter} summarizes the common hyperparameters to be used in the above learning methods.
All the common hyperparameters including for the network architecture are set to the same values as the previous work~\citep{kobayashi2020proximal}.
In all the simulations, the same hyperparameters are used for simplicity and confirmation of robustness for hyperparameters, although the optimal values for each task are different from each other.

\subsection{Selection of comparisons}

\begin{table}[tb]
    \caption{Common hyperparameters for the simulations}
    \label{tab:comparison}
    \centering
    \begin{tabular}{ccc}
        \hline\hline
        Label & Target & $(n, \tau, \nu)$ \\
        \hline
        none & none & $(1, 1.0, \mathrm{inf})$ \\
        hard & all & $(3, 1.0, \mathrm{inf})$ \\
        soft & all & $(1, 0.5, \mathrm{inf})$ \\
        t-soft & all & $(1, 0.5, 1.0)$ \\
        \hline\hline
    \end{tabular}
\end{table}

To verify the benefits of the t-soft update, the comparison methods are selected in this section.
For that purpose, through a toy problem (i.e., CartPoleContinuousBulletEnv-v0), we compare the effects of four factors related to the proposed method: object of the target network (value, policy, all); the update interval of the target network $n = (1, 3, 5)$; the default update rate $\tau \in [0.1, 1]$ with $0.1$ increment; and the degree of freedom for the t-soft update $\nu = (0.1, 1.0, 10.0, \infty)$.
Note that when $\tau = 1$, it is regarded to be the hard update.
The detailed results are described in~\ref{app:select_comp}.

According to the results, four different conditions are selected as the comparisons.
Their hyperparameters, $(n, \tau, \nu)$, is summarized in Table~\ref{tab:comparison}.
Note that all the conditions except ``none'' apply the target network into both the value function and the policy.
The first three correspond to the conventional methods, and the last one is the proposed method.

\begin{figure*}[tb]
    \centering
    \subfigure[Swingup]{
        \includegraphics[keepaspectratio=true,width=0.23\linewidth]{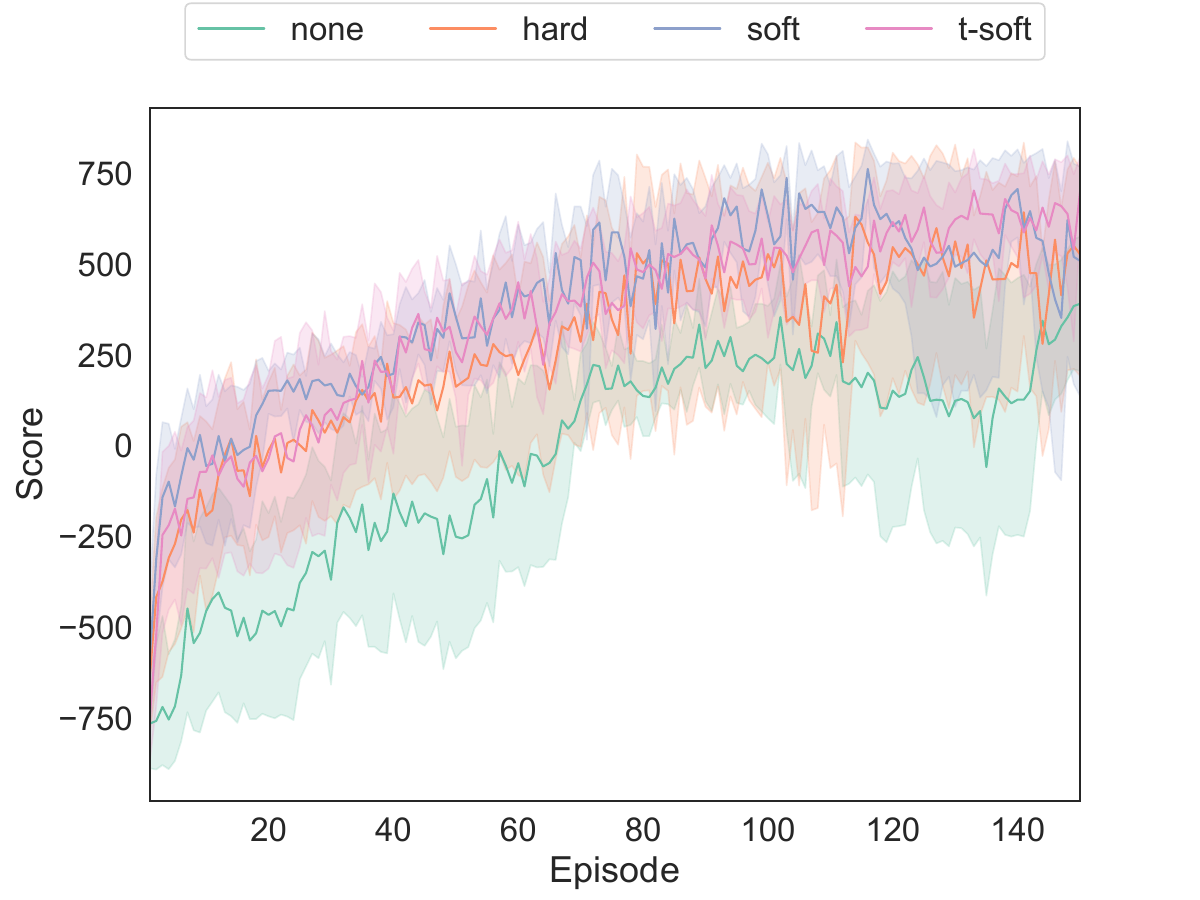}
    }
    \centering
    \subfigure[DoublePendulum]{
        \includegraphics[keepaspectratio=true,width=0.23\linewidth]{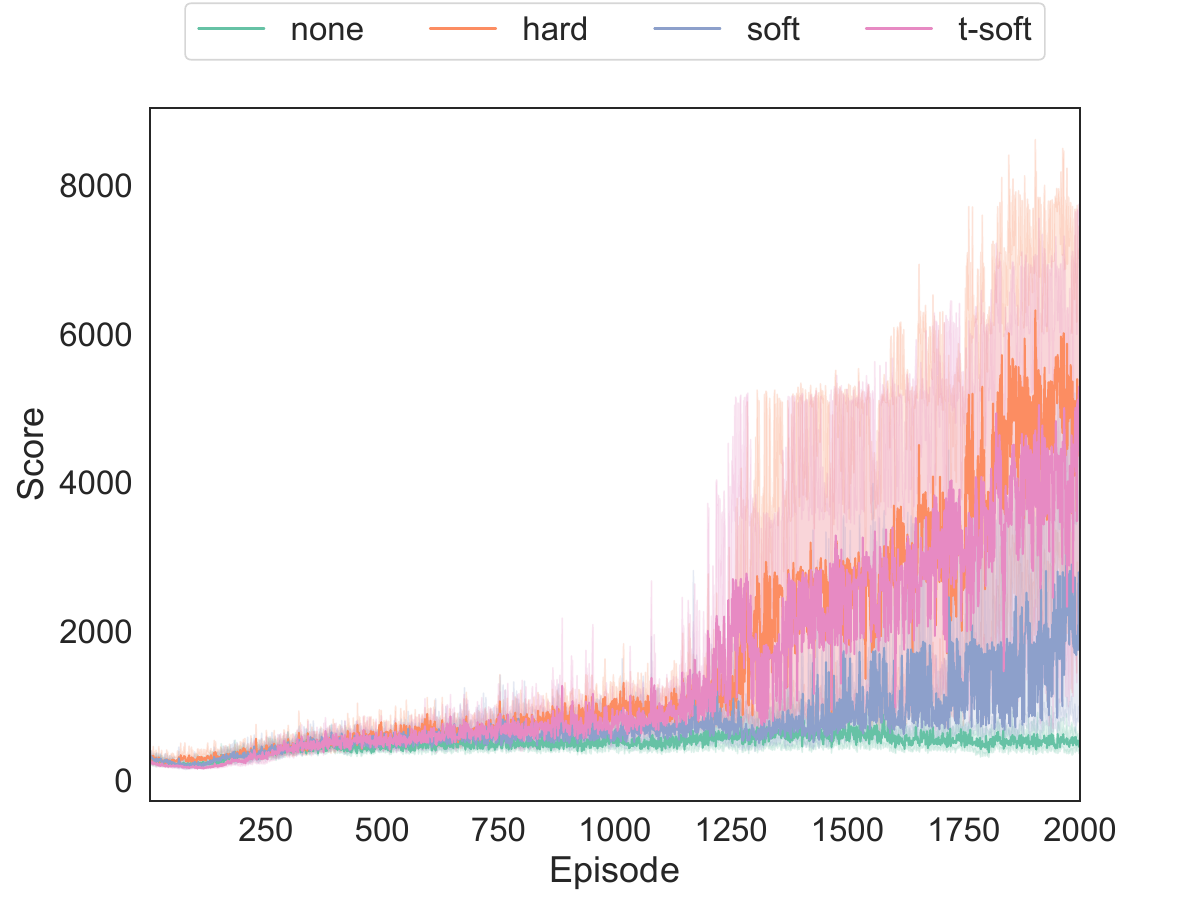}
    }
    \centering
    \subfigure[HalfCheetah]{
        \includegraphics[keepaspectratio=true,width=0.23\linewidth]{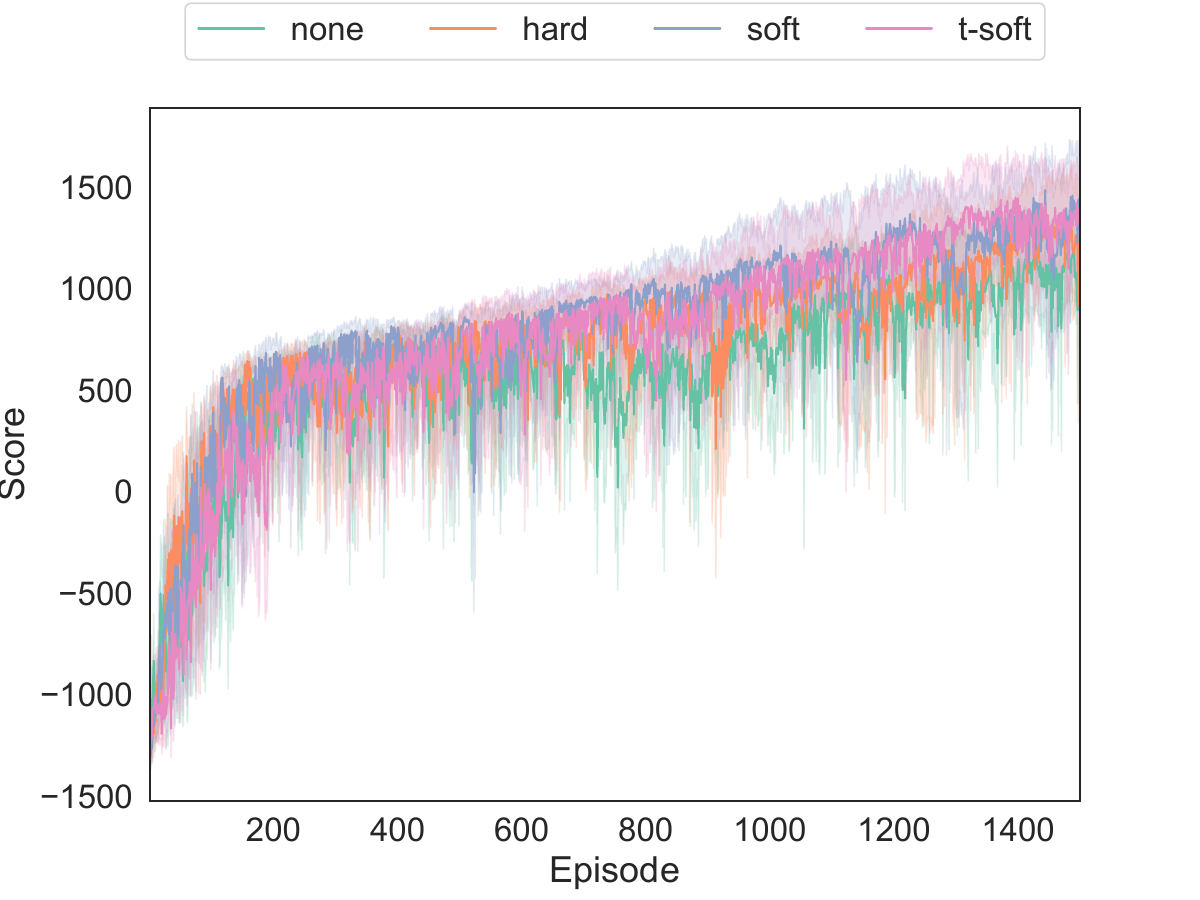}
    }
    \centering
    \subfigure[Ant]{
        \includegraphics[keepaspectratio=true,width=0.23\linewidth]{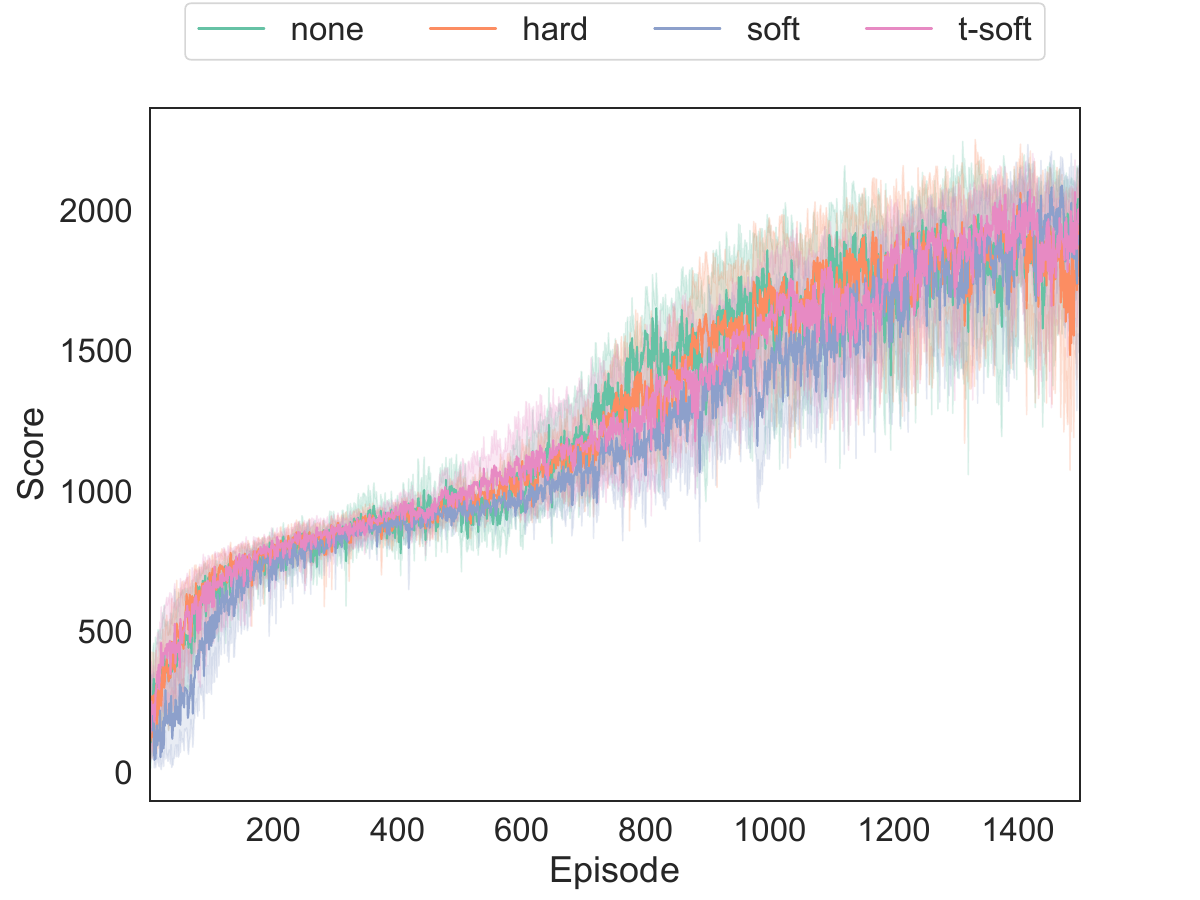}
    }
    \caption{Learning curves of four benchmark tasks:
        the sum of rewards at each episode are given as the score;
        the corresponding shaded areas show the 95~\% confidence intervals;
        without the target network, the learning performance in any task was deteriorated in comparison with others;
        the hard update might get overfitting in the Ant task, and the soft update in the DoublePendulum task failed to learn the optimal policy in most of the trials;
        in contrast, the t-soft update achieved stable learning of all the tasks.
    }
    \label{fig:sim_learn}
\end{figure*}

\begin{figure}[tb]
    \centering
    \includegraphics[keepaspectratio=true,width=0.98\linewidth]{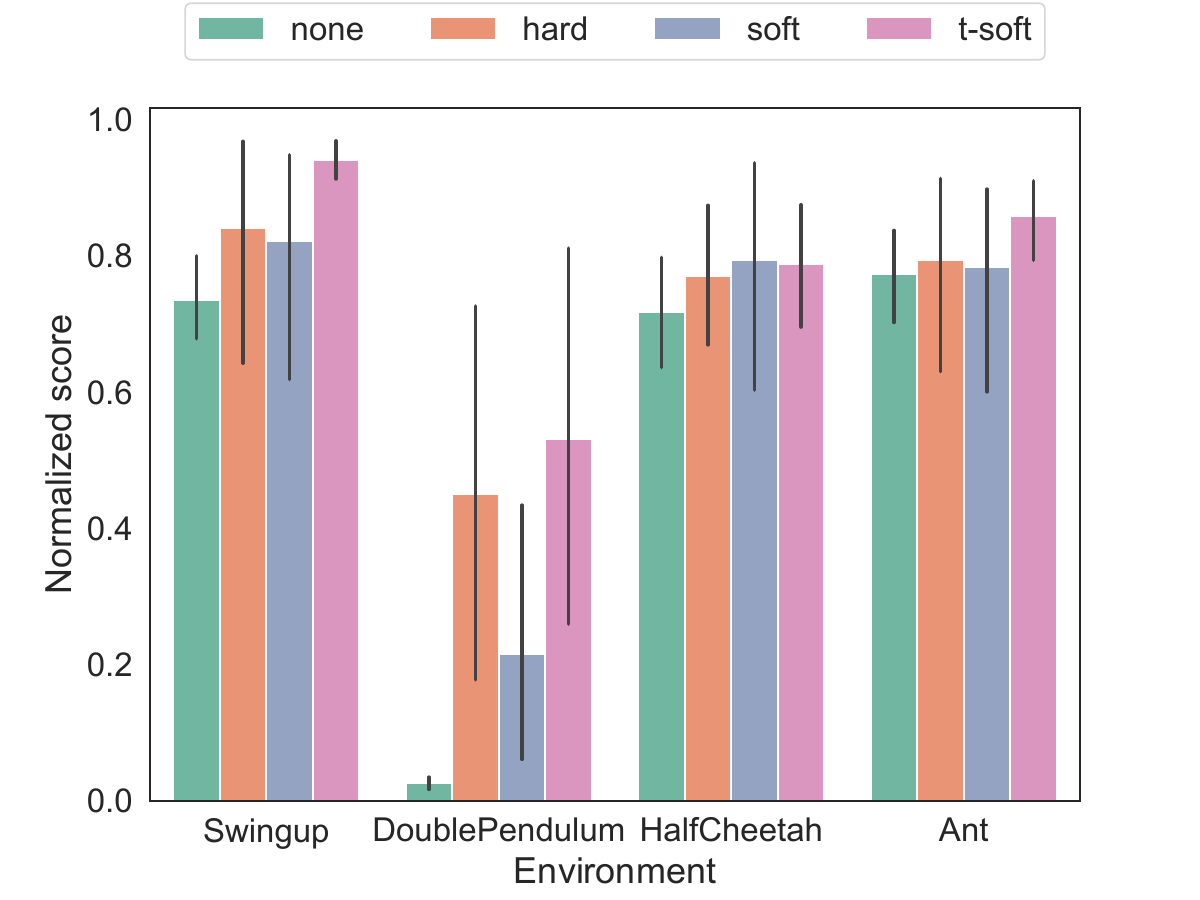}
    \caption{Summary of four benchmark tasks as bar plots:
        For clear visualization, the raw scores were normalized by maximum and minimum scores in each task;
        the 95~\% confidence intervals were shown as black line segments on the respective bars;
        the proposed t-soft update outperformed the other methods in all the tasks except the HalfCheetah task, where there was no significant difference when using the target network;
        the t-soft update remarkably reduced the variance of the learning results.
    }
    \label{fig:sim_summary}
\end{figure}

\begin{figure*}[tb]
    \centering
    \subfigure[$\bar{\tau}_i$ on Swingup]{
        \includegraphics[keepaspectratio=true,width=0.23\linewidth]{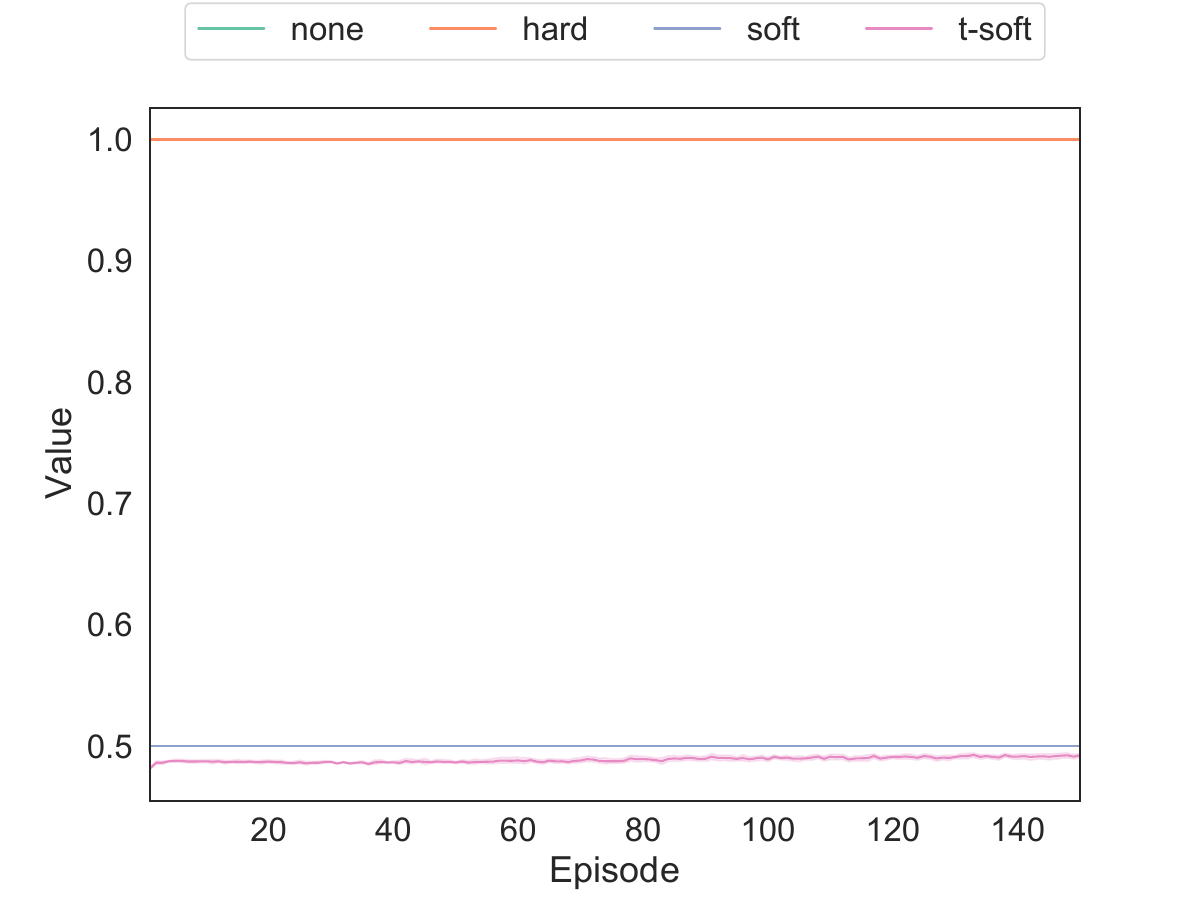}
    }
    \centering
    \subfigure[$\bar{\tau}_i$ on DoublePendulum]{
        \includegraphics[keepaspectratio=true,width=0.23\linewidth]{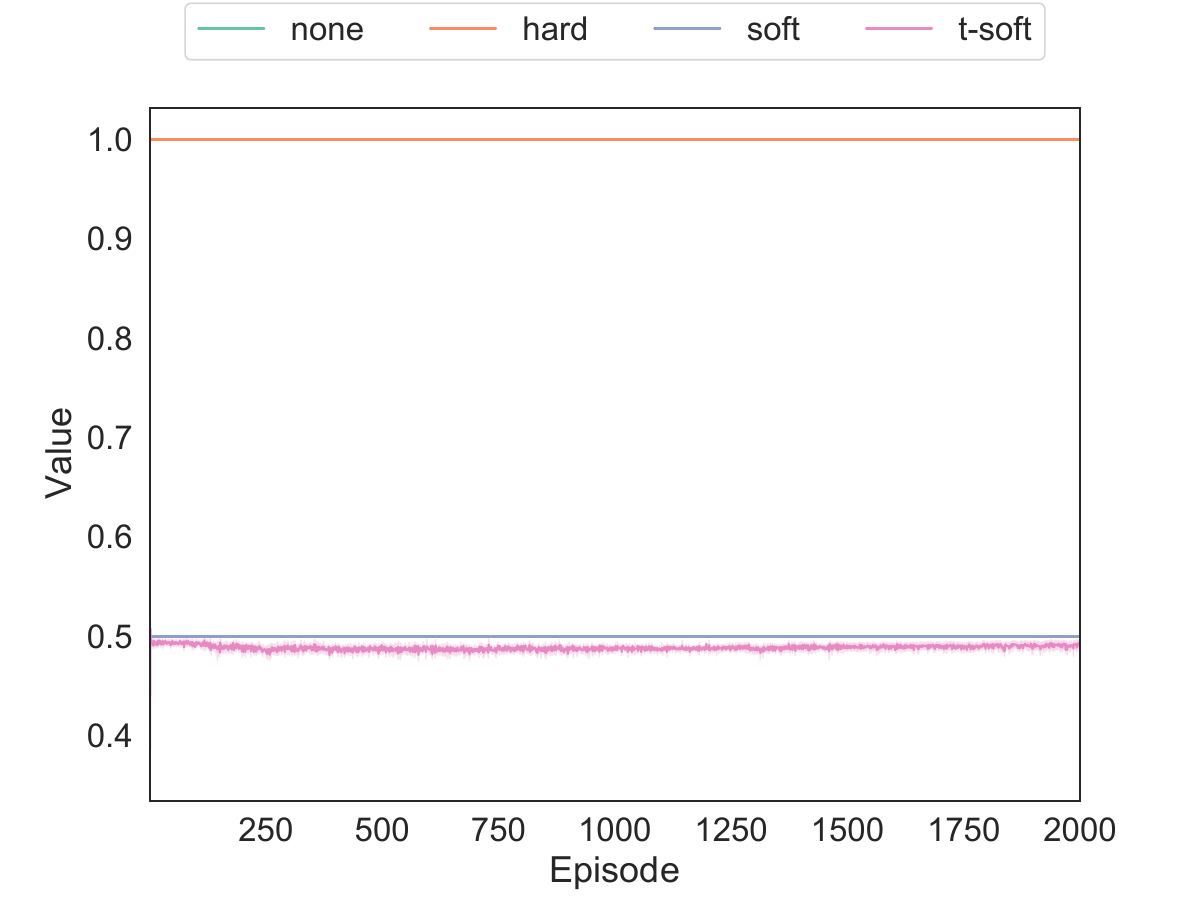}
    }
    \centering
    \subfigure[$\bar{\tau}_i$ on HalfCheetah]{
        \includegraphics[keepaspectratio=true,width=0.23\linewidth]{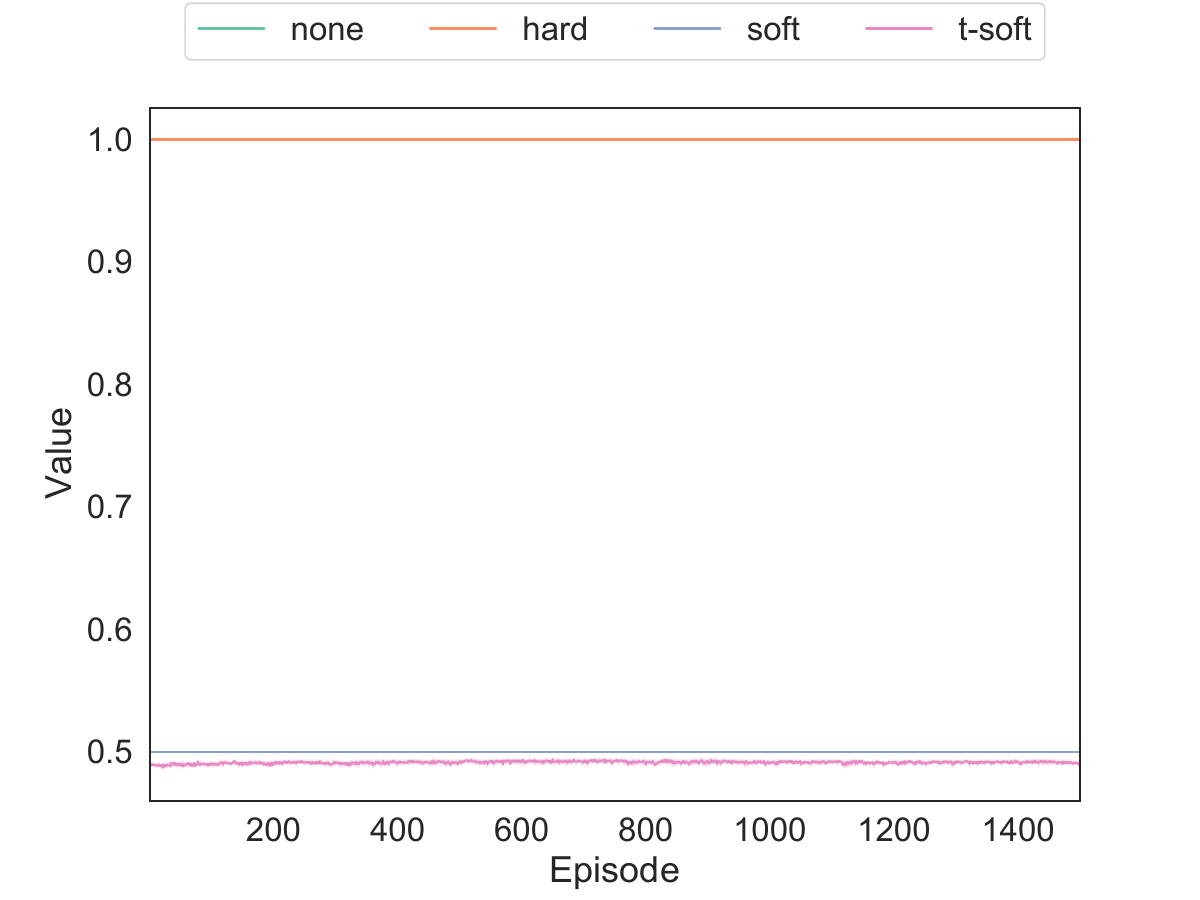}
    }
    \centering
    \subfigure[$\bar{\tau}_i$ on Ant]{
        \includegraphics[keepaspectratio=true,width=0.23\linewidth]{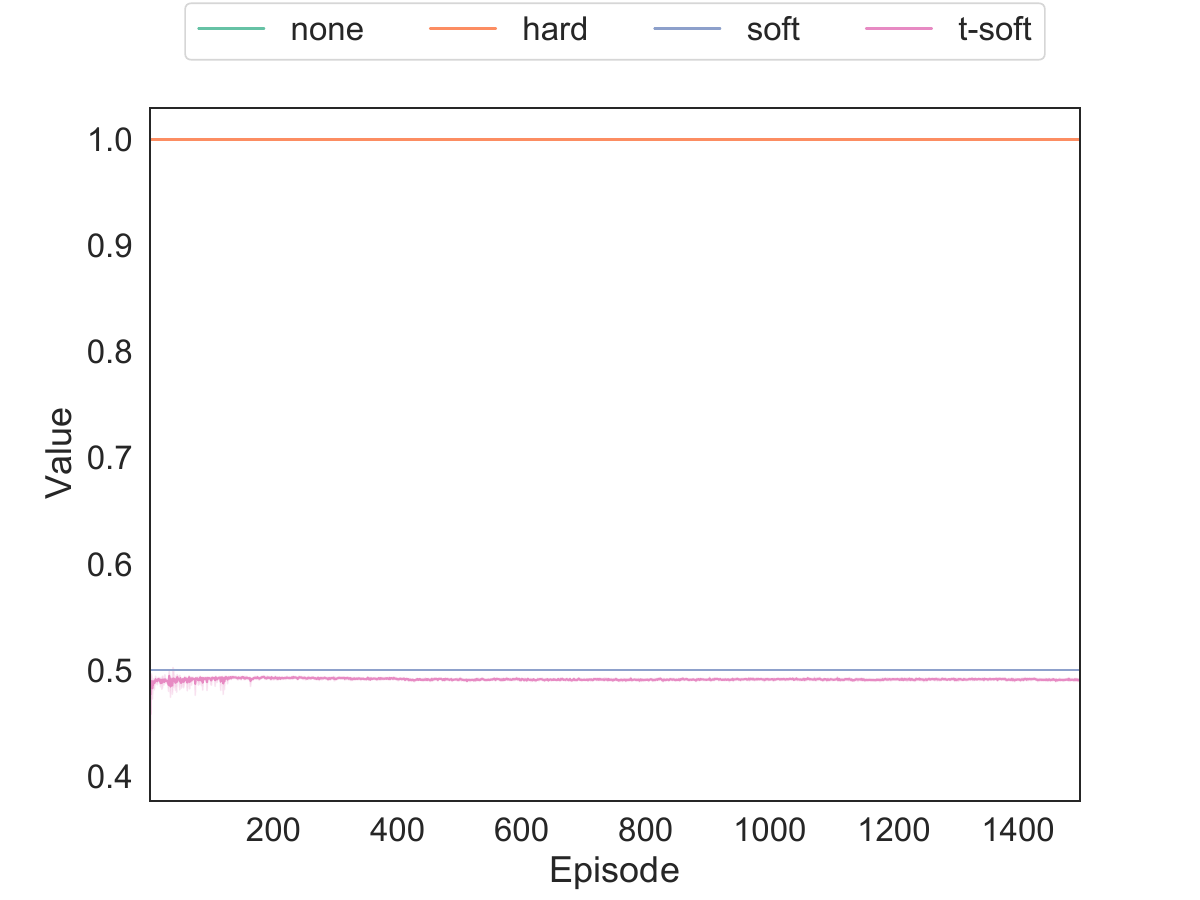}
    }
    \centering
    \subfigure[$\max(\tau_i)$ on Swingup]{
        \includegraphics[keepaspectratio=true,width=0.23\linewidth]{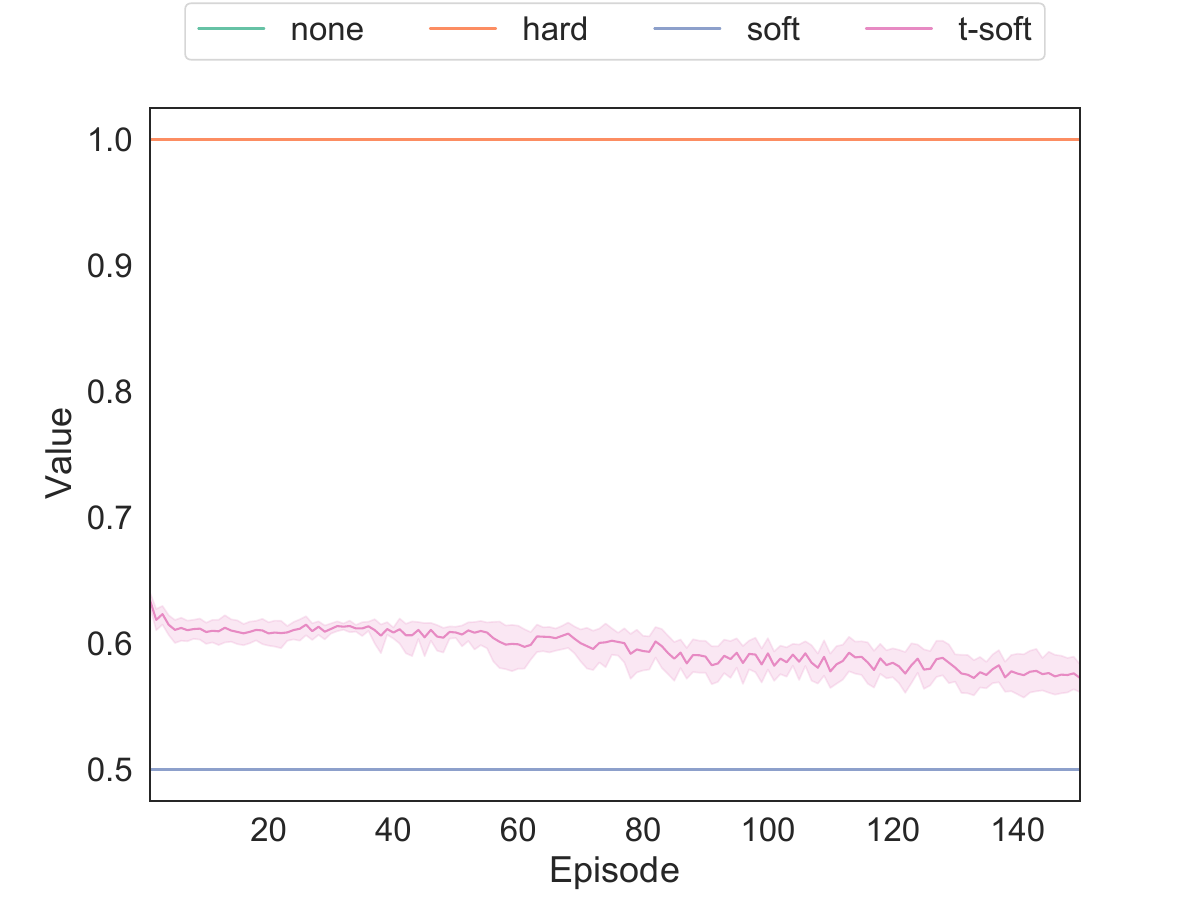}
    }
    \centering
    \subfigure[$\max(\tau_i)$ on DoublePendulum]{
        \includegraphics[keepaspectratio=true,width=0.23\linewidth]{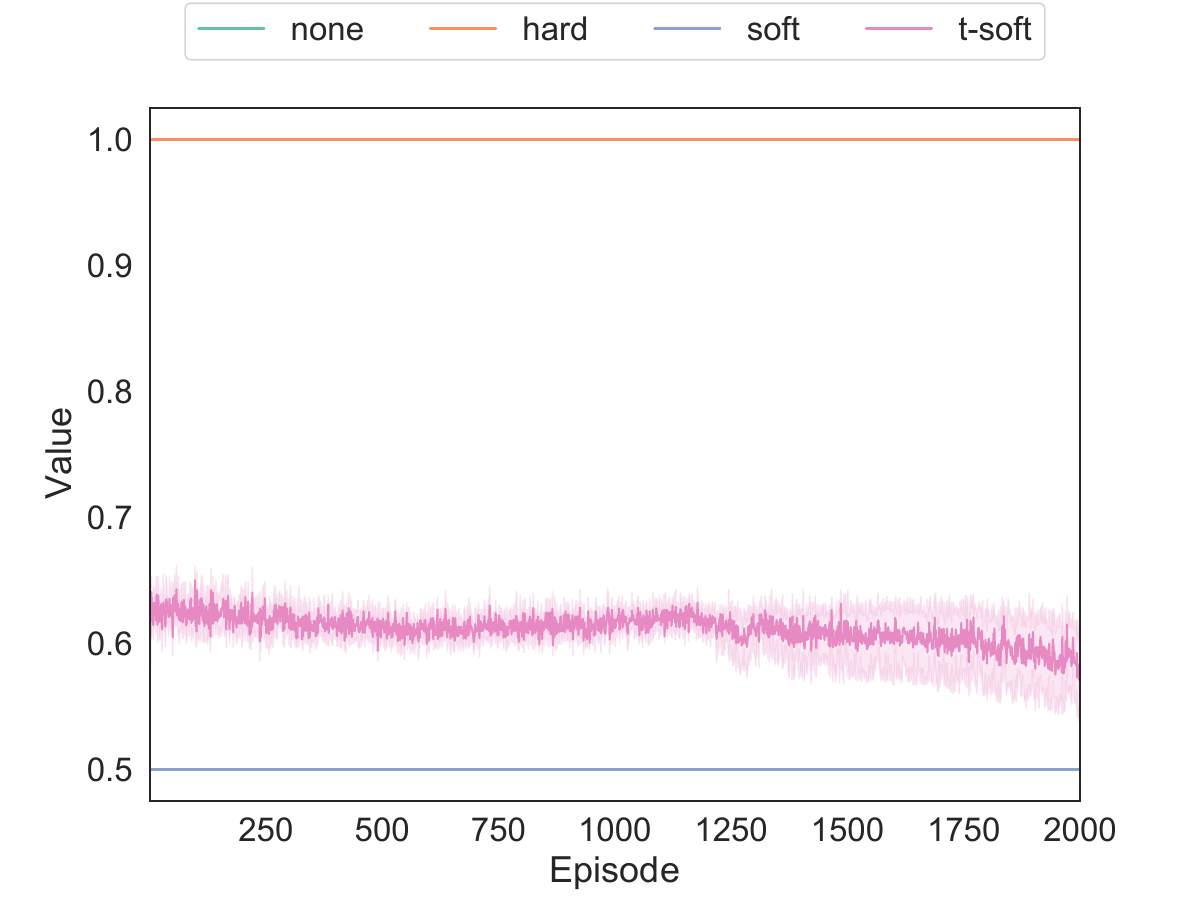}
    }
    \centering
    \subfigure[$\max(\tau_i)$ on HalfCheetah]{
        \includegraphics[keepaspectratio=true,width=0.23\linewidth]{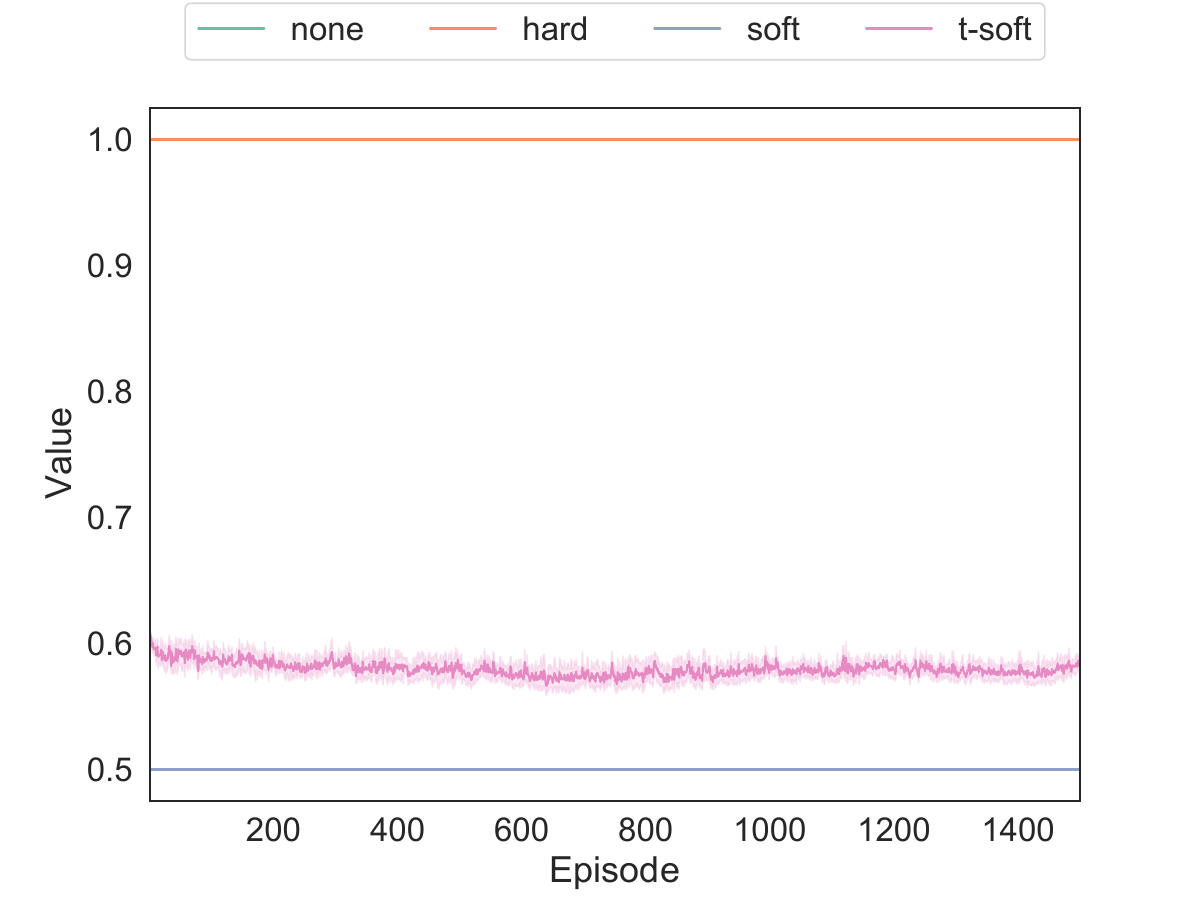}
    }
    \centering
    \subfigure[$\max(\tau_i)$ on Ant]{
        \includegraphics[keepaspectratio=true,width=0.23\linewidth]{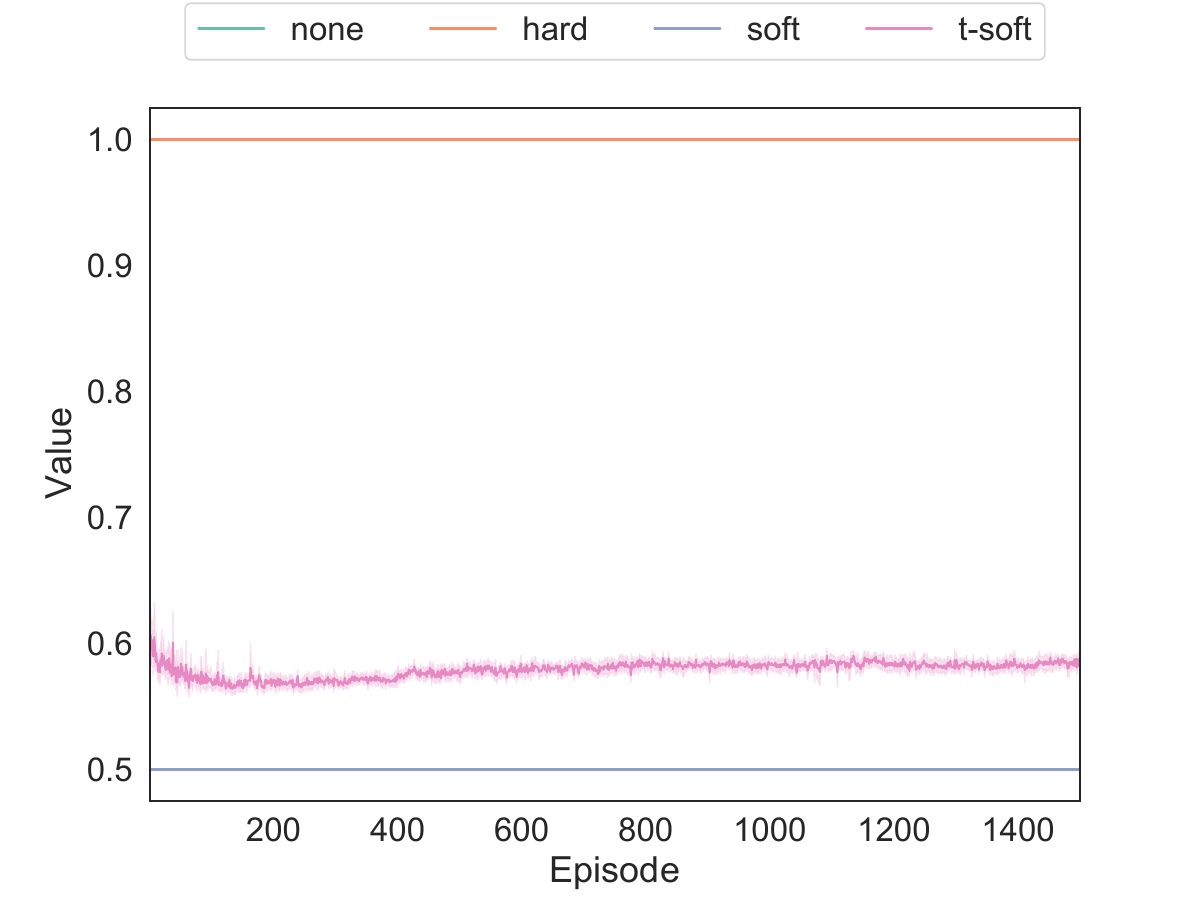}
    }
    \centering
    \subfigure[$\min(\tau_i)$ on Swingup]{
        \includegraphics[keepaspectratio=true,width=0.23\linewidth]{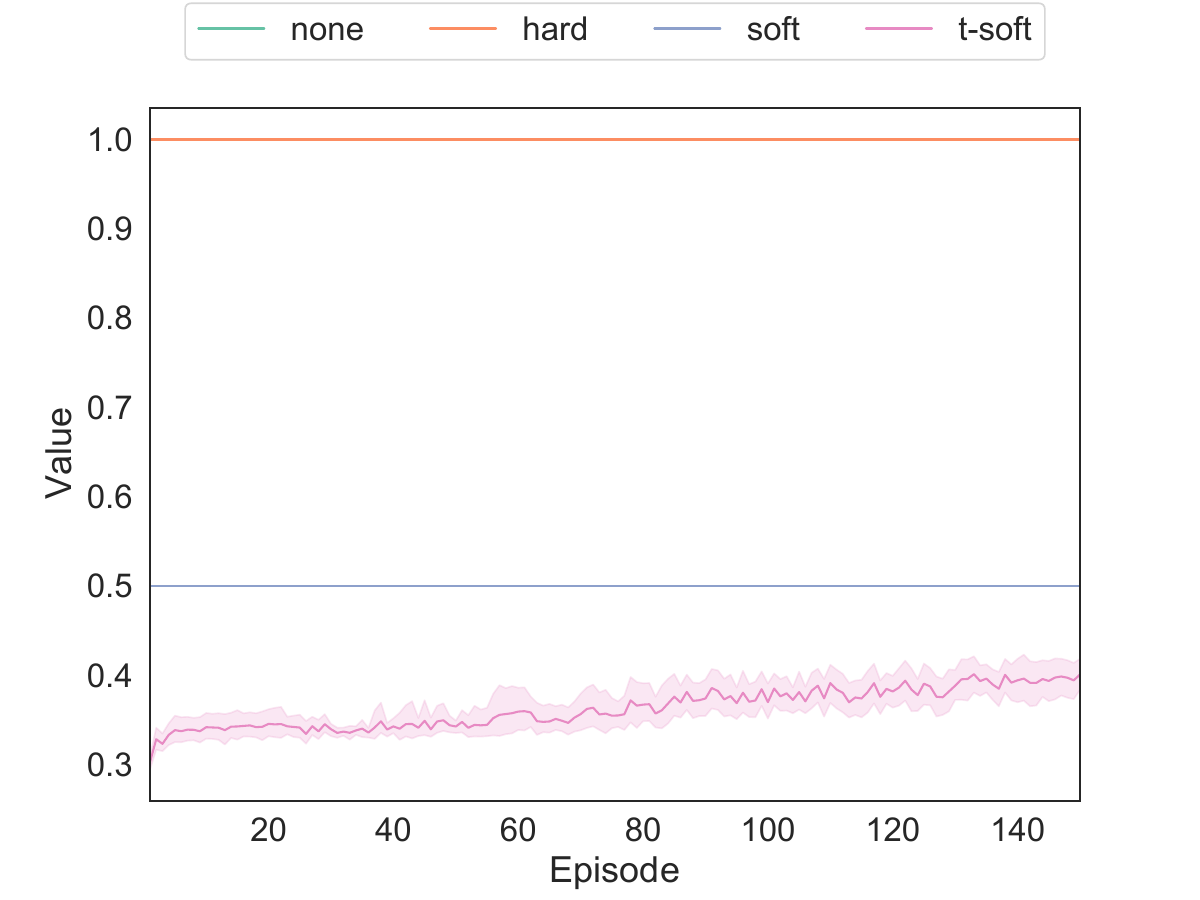}
    }
    \centering
    \subfigure[$\min(\tau_i)$ on DoublePendulum]{
        \includegraphics[keepaspectratio=true,width=0.23\linewidth]{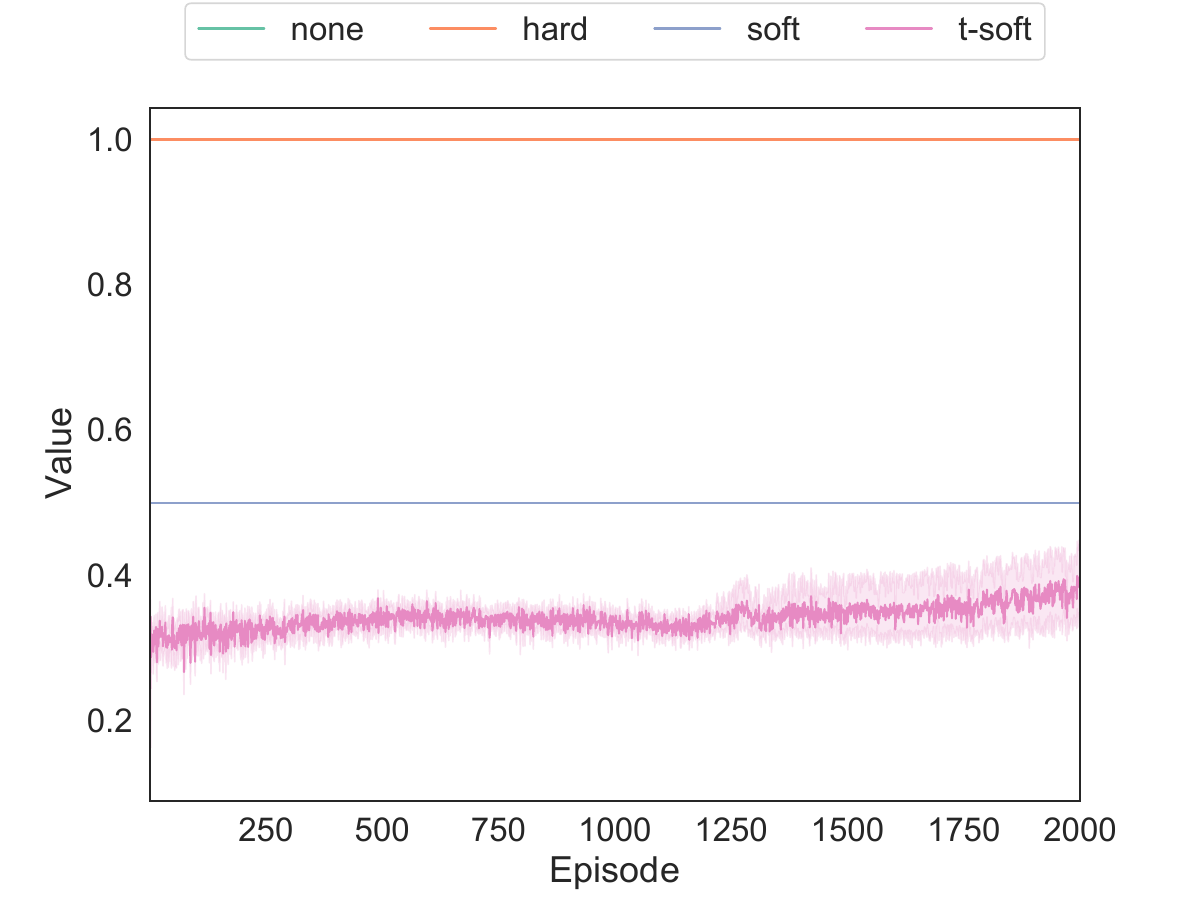}
    }
    \centering
    \subfigure[$\min(\tau_i)$ on HalfCheetah]{
        \includegraphics[keepaspectratio=true,width=0.23\linewidth]{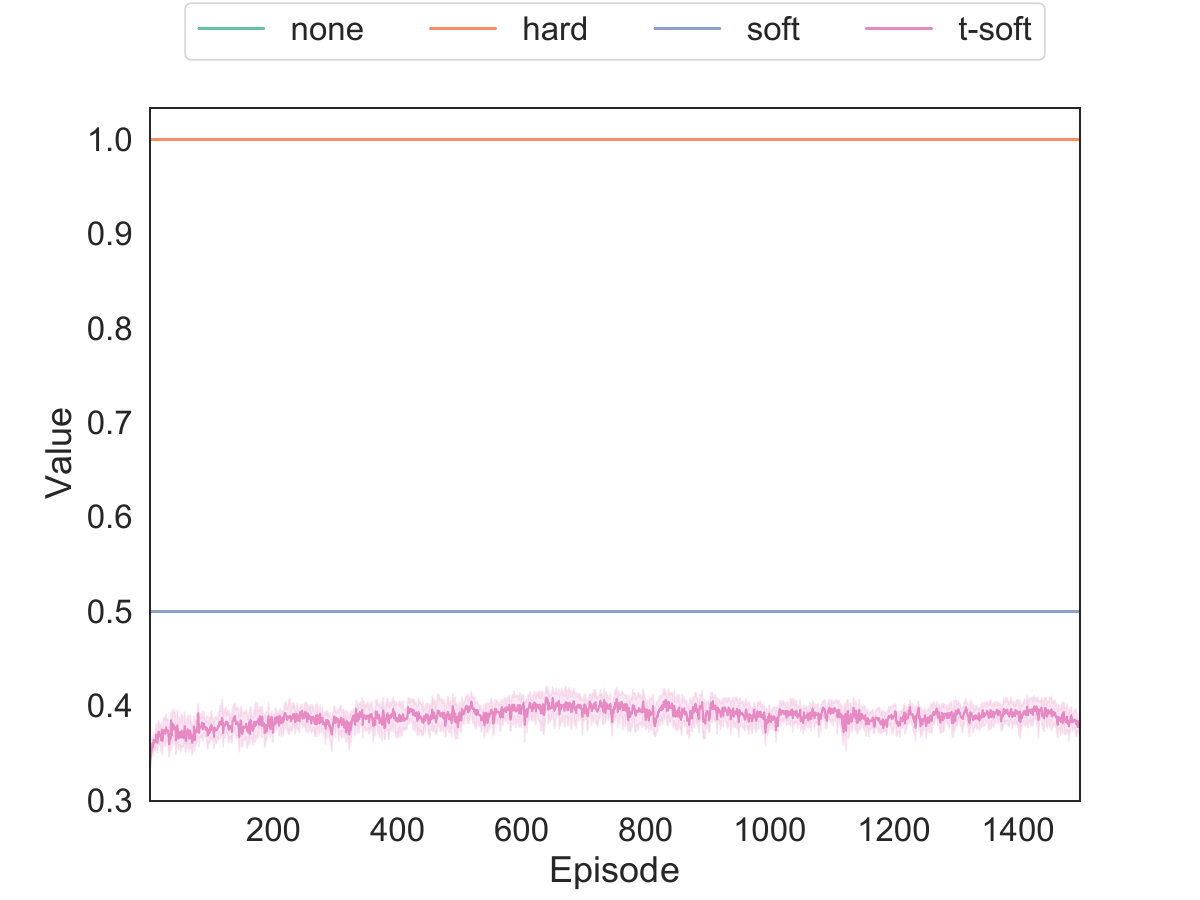}
    }
    \centering
    \subfigure[$\min(\tau_i)$ on Ant]{
        \includegraphics[keepaspectratio=true,width=0.23\linewidth]{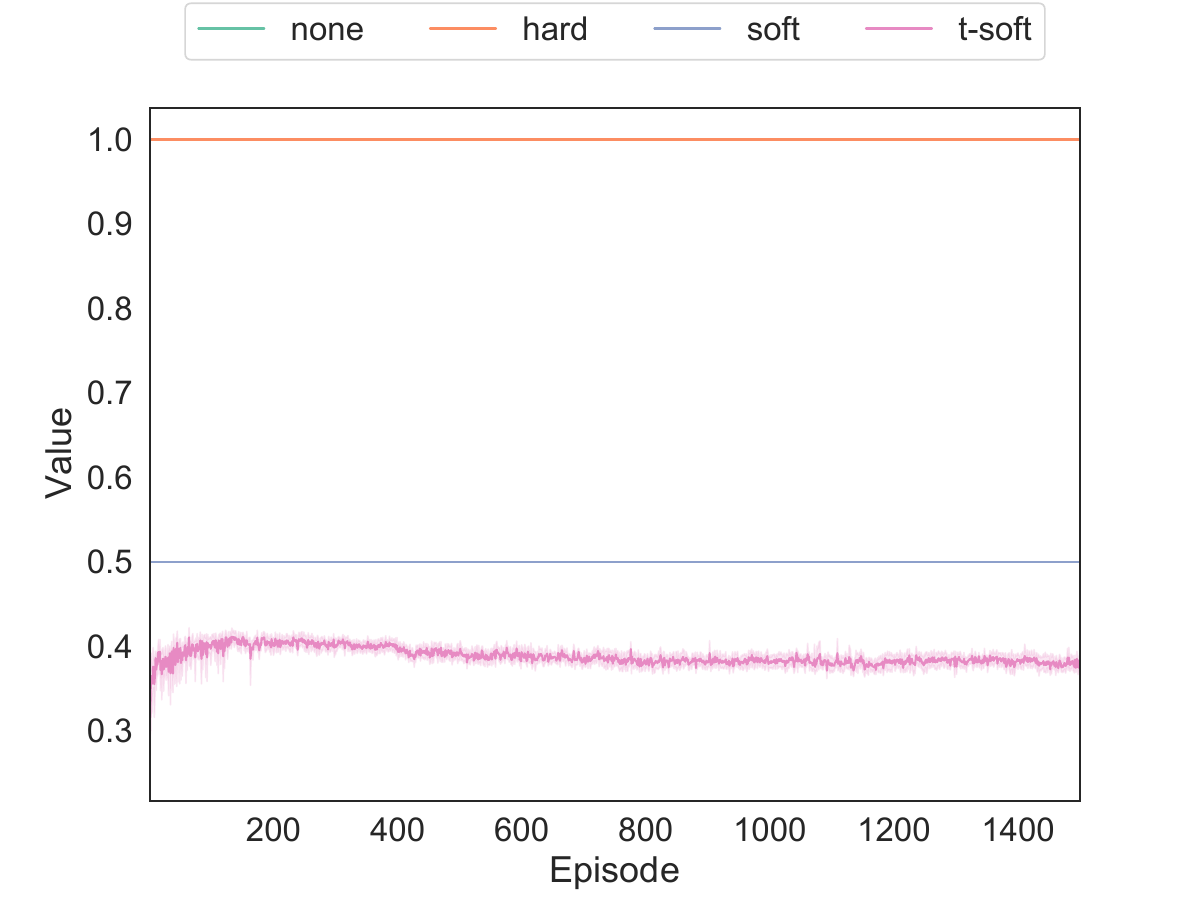}
    }
    \caption{Adaptive behavior of the t-soft update represented by $\tau_i$:
        $tau_i$ is given for each subset of parameters, and to clearly visualize them statistically, their mean $\bar{\tau}_i$, maximum $\max(\tau_i)$, and minimum $\min(\tau_i)$ were depicted as their learning curves;
        in (a)--(d), the t-soft update made $\bar{\tau}_i$ slightly smaller than $\tau$ overall, that is, its macro behavior converges to the traditional soft update while slightly suppressing the updates;
        in (k)--(l), $\max(\tau_i)$ and $\min(\tau_i)$ were far away from $\tau$ about $0.1$, that is, the micro behavior of the t-soft update can be divided into accelerated and suppressed updates for the respective subsets.
    }
    \label{fig:stat_tar}
\end{figure*}

\begin{figure}[tb]
    \centering
    \subfigure[Swingup]{
        \includegraphics[keepaspectratio=true,width=0.465\linewidth]{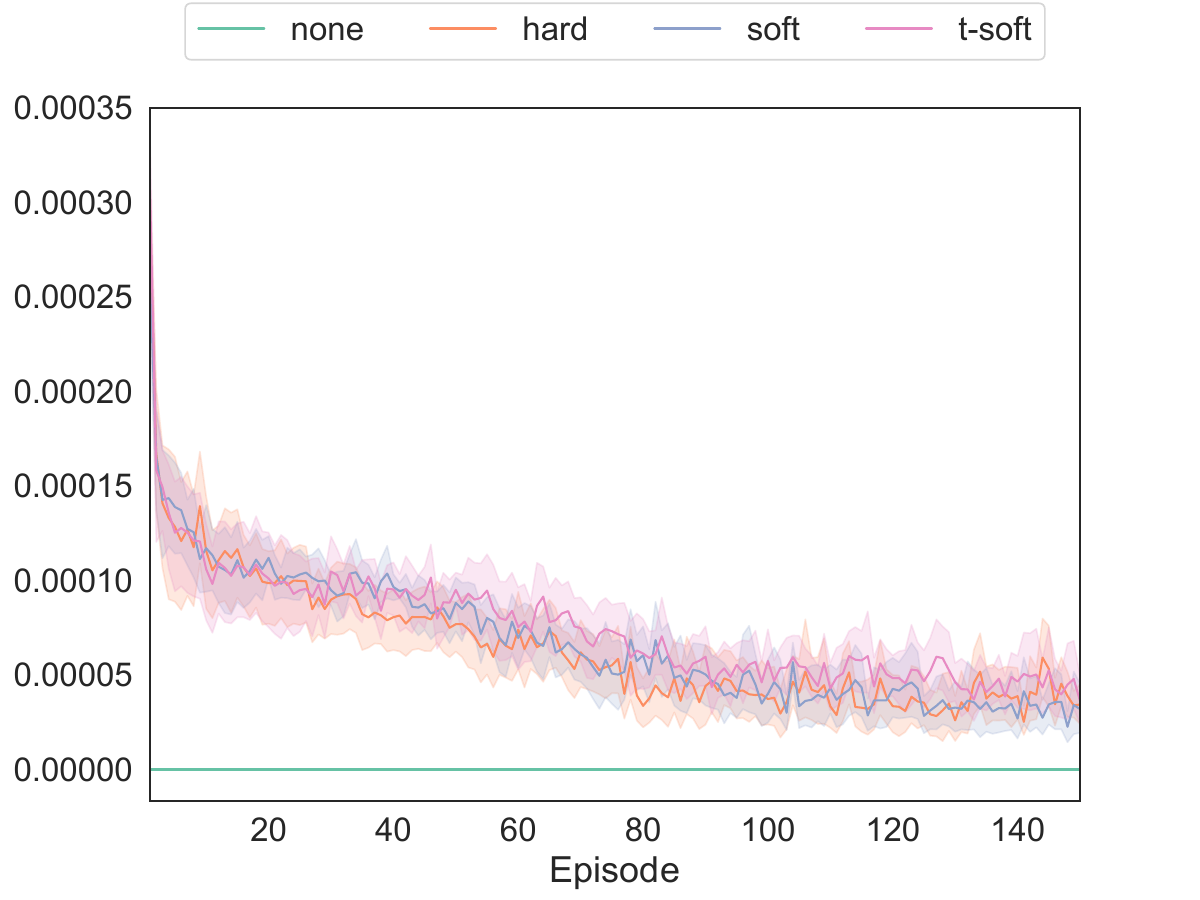}
    }
    \centering
    \subfigure[HalfCheetah]{
        \includegraphics[keepaspectratio=true,width=0.465\linewidth]{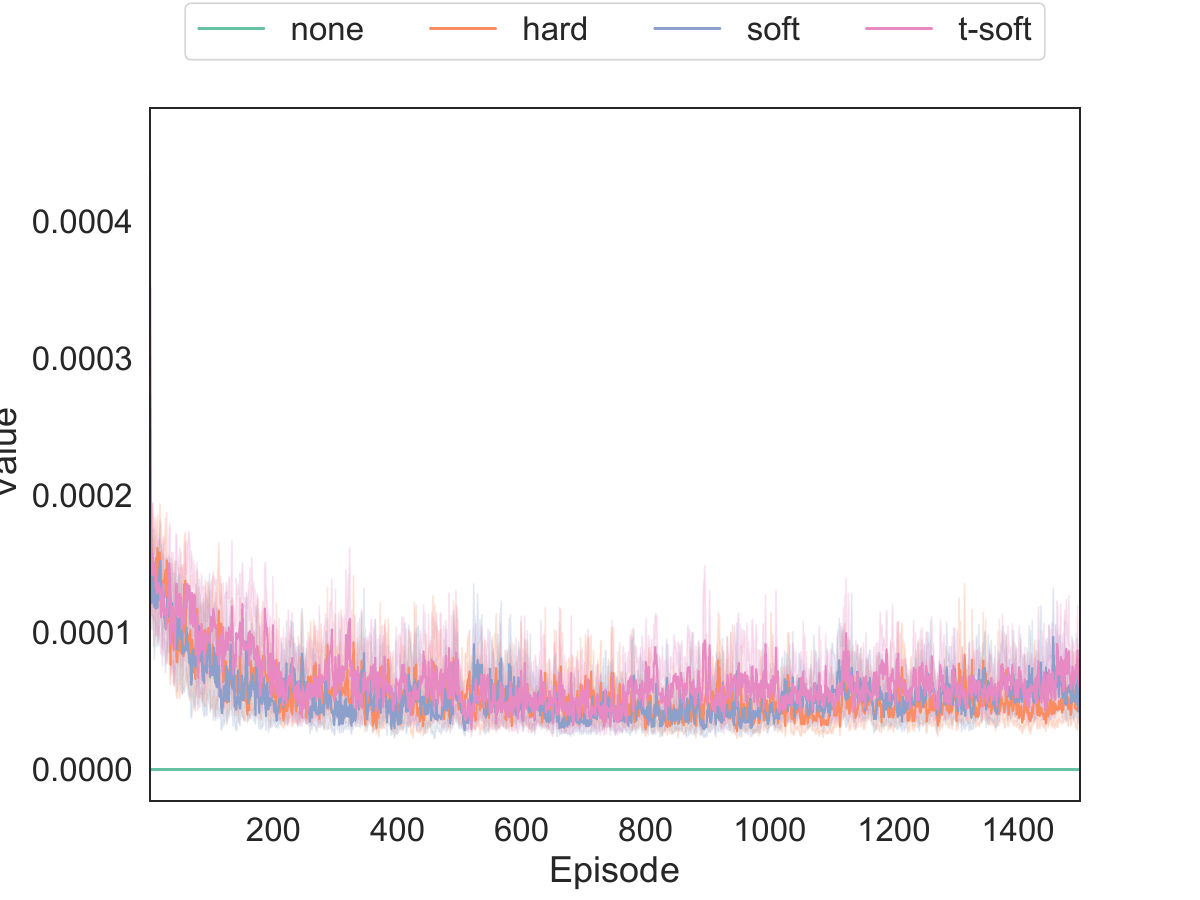}
    }
    \caption{Typical examples of the divergence between the main and target networks:
        as a natural result of the fact that about 90\% of the parameters are updated in three soft updates with $\tau \simeq 0.5$, the hard and soft updates obtained the similar divergences;
        in contrast, with the t-soft update, the divergence became slightly far from the others probably because of the robust updates for incorrect updates of the main network.
    }
    \label{fig:sim_tar_diff}
\end{figure}

\subsection{Results}

With the comparisons selected in the above, 10 trials with different random seeds were performed for each condition.
Learning curves for them are illustrated in Fig.~\ref{fig:sim_learn}.
The test results after learning are also depicted in Fig.~\ref{fig:sim_summary}.

Consistent with the previous reports, we found that the case without the target network failed learning any task frequently.
As can be seen in Fig.~\ref{fig:sim_learn}(b) (i.e., the DoublePendulum task), the soft update often failed to find a global optimum, while the hard and t-soft updates found it.
On the other hand, the hard update caused overfitting finally in Fig.~\ref{fig:sim_learn}(d) (i.e., the Ant task), while the soft and t-soft updates did not so.
Namely, the proposed t-soft update achieved stable learning for any tasks, in contrast to the traditional methods, which resulted in unstable learning for some tasks.

Indeed, the t-soft update outperformed the others in the test results as shown in Fig.~\ref{fig:sim_summary}.
It remarkably reduced the variance of learning results represented by 95\% confidence intervals as black line segments.
Note that the variance of DoublePendulum task is higher than that of the other tasks since the score on success is much larger than on failure.
In addition, there seems to be no significant difference between the soft and t-soft updates in the HalfCheetah task.
This may be because the learning method used in these simulations lacked the capability to acquire better policies and was not effective in suppressing wrong updates to emphasize correct updates.

To investigate the internal behaviors in the t-soft update, Fig.~\ref{fig:stat_tar} illustrates $\tau_i$ in eq.~\eqref{eq:tsoft_upd} during learning.
Note that mean, maximum, and minimum are computed as its statistics since $\tau_i$ is given for each subset of the parameters set $\theta$ (and $\phi$).
From Figs.~\ref{fig:stat_tar}(a)--(c) for the mean of $\tau_i$, $\bar{\tau}_i$, we can see that $\bar{\tau}_i$ is close but slightly smaller to/than the designed value $\tau$.
This suggests that from a macro perspective, the t-soft update is consistent with the soft update, thereby avoiding extremely delayed or accelerated updates.
In contrast, Figs.~\ref{fig:stat_tar}(e)--(l) for the maximum and minimum of $\tau_i$, $\max(\tau_i)$ and $\min(\tau_i)$ respectively, indicate a micro perspective for the respective subsets.
Specifically, some of subsets obtains $\tau_i > \tau$, which accelerates the updates toward the main network, and the remainder (i.e., $\tau_i < \tau$) would be conservatively updated.
Thus, from a micro perspective, we confirmed the adaptive behavior of adjusting the amount of updates (i.e., the acceleration or suppression) for each subset.

As examples to show the difference of the update rule of target network, Fig.~\ref{fig:sim_tar_diff} is drawn as the divergence of parameters for the main networks $\theta$ and $\eta$ and the target networks $\phi$ and $\zeta$, $|\theta - \phi| + |\eta - \zeta|$.
First, the divergences obtained the hard and soft updates were mostly overlapped since the soft update with $\tau = 0.5$ would complete roughly 90\% of its update in 3 times.
The divergence obtained by the t-soft update was also mostly overlapped, but tends to be slightly larger than the others.
This is a result of $\tau_i$ being slightly smaller than $\tau$.
However, it is important to note that the t-soft update caught up with the other updates sometimes, such as at the beginning of Fig.~\ref{fig:sim_tar_diff}(a) and in the middle (around 700 episode) of Fig.~\ref{fig:sim_tar_diff}(b).
That is, the necessary updates were properly done.

\section{Conclusion}
\label{sec:conclusion}

\subsection{Summary}

This paper proposed a new robust update rule of the target network for DRL, so-called the t-soft update.
By focusing on the fact that the conventional soft update is based on the EMA, which can be replaced with the noise-robust version derived from the student-t distribution, we designed the proposed t-soft update.
In practice, its computation and memory costs were reduced by assuming a one-dimensional model for each subset of parameters.
The behavior of t-soft update can be distinguished into two types based on the parameters of the new main network and the amount of deviation from the previous experiences.
Specifically, if significant deviations from the previous experiences are observed, careful updates are chosen by restraining the updates.
Otherwise, accelerated updating will mitigate slowdown of learning speed, as concerned in the conventional soft update.
In PyBullet robotics simulations, the actor-critic algorithm with the t-soft update outperformed the conventional methods (i.e., the method without target network; the hard update; and the soft update) in terms of the sum of rewards after learning and/or its variance.
Analysis of the training process showed that, as expected, the t-soft update is globally consistent with the standard soft update, and the update rates are locally (in each subset) adjusted for acceleration or suppression.

\subsection{Discussion and future work}

As the learning method for the simulations, online-learning-based method was employed.
However, in the recent DRL trends, the experience replay~\citep{lin1992self,andrychowicz2017hindsight} is the major technique to make learning sample-efficient.
This has also been studied in neuroscience, reporting that valuable experiences are replayed and organized in the brain, especially during sleep~\citep{singer2009rewarded,gulati2017neural}.
In addition, it has been suggested that not only skill level but also learning speed would be increased after sleep~\citep{walker2003sleep,kuriyama2004sleep}.
This suggestion about the increase of learning speed gives us the possibility that the target network is directly updated by the experience replay to generate more optimal targets, although the current method is for updating the main network.

Therefore, in future work, we will investigate the use of experience replay for updating the target network directly.
In this way, the target network, which has been only following the main network, is expected to present the more appropriate targets and accelerate learning.
In addition, the best algorithm with the t-soft update will be applied to complicated robotic tasks.

\section*{Acknowledgements}
This work was supported by JSPS KAKENHI, Grant-in-Aid for Scientific Research (B), Grant Number 20H04265.

\appendix
\section{Actor-critic algorithm}
\label{app:actor_critic}

Let us briefly introduce the actor-critic algorithm~\citep{williams1992simple,peters2008natural}.
To explicitly optimize the policy parameterized by $\eta$, the following loss function is minimized.
\begin{align}
    \mathcal{L}(\eta) &= - \mathbb{E}_{a_t \sim \pi(\eta), s_t \sim p_T}[A(s_t, a_t; \theta, \phi)]
\end{align}
where $A(s_t, a_t; \theta, \phi) = Q(s_t, a_t; \phi) - V(s_t; \theta) = r_t + \gamma V(s_{t+1}; \phi) - V(s_t; \theta)$.
The gradient of this loss function is derived as follows:
\begin{align}
    \nabla_\eta \mathcal{L}(\eta) &= - \mathbb{E}_{s_t \sim p_T} \Biggl [ \int A(s_t, a_t; \theta, \phi) \nabla_\eta \pi(a_t \mid s_t; \eta) da_t \Biggr ]
    \nonumber \\
    &= - \mathbb{E}_{s_t \sim p_T} \Biggl [ \int \pi(a_t \mid s_t; \eta) A(s_t, a_t; \theta, \phi)
    \nonumber \\
    &\times \cfrac{\nabla_\eta \pi(a_t \mid s_t; \eta)}{\pi(a_t \mid s_t; \eta)} da_t \Biggr ]
    \nonumber \\
    &= - \mathbb{E}_{a_t \sim \pi(\eta), s_t \sim p_T}[A(s_t, a_t; \theta, \phi) \nabla_\eta \ln \pi(a_t \mid s_t; \eta)]
\end{align}
Here, the expectation is approximated by Monte Calro method with numerous samples.
The above gradient is used for the SGD optimizer.
Note that, by using importance sampling, the baseline policy with the parameters set $\zeta$ can be introduced in this loss function as follows:
\begin{align}
    \nabla_\eta \mathcal{L}(\eta) &= - \mathbb{E}_{a_t \sim \pi(\zeta), s_t \sim p_T} \Biggl [ \cfrac{\pi(a_t \mid s_t; \eta)}{\pi(a_t \mid s_t; \zeta)}
    \nonumber \\
    &\times A(s_t, a_t; \theta, \phi) \nabla_\eta \ln \pi(a_t \mid s_t; \eta) \Biggr ]
\end{align}
That is, actions are sampled from the baseline policy $\zeta$, which is slowly and stably updated towards the main policy $\eta$.
As well as the case of the value function, this update of $\zeta$ is performed by the (t-)soft update.

\section{Derivation of update rule for scale parameter}
\label{app:student_sigma}

The scale parameter $\sigma \in \mathbb{R}_+^d$ in $d$-dimensional diagonal student-t distribution with the degree of freedom $\nu \in \mathbb{R}_+$ is updated according to the observed deviation $\Delta \in \mathbb{R}^d$ and the gradient ascent for maximum likelihood.
The probability density function of this distribution and its log likelihood are given as follows:
\begin{align}
    p(\Delta) &= Z \left ( \prod_{k=1}^d \sigma_k^2 \right )^{-\cfrac{1}{2}} \left ( 1 + \cfrac{1}{\nu} \sum_{k=1}^d \cfrac{\Delta_k^2}{\sigma_k^2} \right )^{-\cfrac{\nu + d}{2}}
    \nonumber \\
    \ln p(\Delta) &= \ln Z - \cfrac{1}{2} \sum_{k=1}^d \ln \sigma_k^2 - \cfrac{\nu + d}{2} \ln \left ( 1 + \cfrac{1}{\nu} D \right )
\end{align}
where $Z$ denotes the normalization factor for satisfying $\int p(\Delta) d\Delta = 1$.
In addition, $D = \sum_{k=1}^d \Delta_k^2 / \sigma_k^2$ is substituted for simplicity.

To maximize the above log likelihood w.r.t $\sigma_k^2$, the gradient is derived as follows:
\begin{align}
    \nabla_{\sigma_k^2} \ln p(\Delta) &= - \cfrac{1}{2} \cfrac{1}{\sigma_k^2} - \cfrac{\nu + d}{2} \cfrac{- \Delta_k^2 / \sigma_k^4}{\nu + D}
    \nonumber \\
    &= \cfrac{1}{2\sigma_k^4} \left ( \cfrac{(\nu + d)\Delta_k^2 - (\nu + D)\sigma_k^2}{\nu + D} \right )
    \nonumber \\
    &= \cfrac{1}{2\sigma_k^4} \left ( \cfrac{\nu(\Delta_k^2 - \sigma_k^2) + d \Delta_k^2 - D \sigma_k^2}{\nu + D} \right )
\end{align}
Here, if $d=1$ like in the implementation of the t-soft update, $d \Delta_k^2 - D \sigma_k^2 = \Delta_k^2 - \Delta_k^2 \sigma_k^{-2} \sigma_k^2 = 0$.
That is, in our case (i.e., $\sigma_k = \sigma_i$, $\Delta_k = \Delta_i$, and $(\nu + 1)/(\nu + D) = w_i$), the above gradient can be derived as follows:
\begin{align}
    \nabla_{\sigma_i^2} \ln p(\Delta_i) = \cfrac{1}{2\sigma_i^4} \cfrac{\nu}{\nu + 1} w_i (\Delta_i^2 - \sigma_i^2)
\end{align}

With the adaptive step size for the gradient ascent, $2\sigma_i^4 \tau$, the update rule of $\sigma_i^2$ is finally consistent with eq.~\eqref{eq:tsoft_var}.
\begin{align}
    \sigma_i^2 &\gets \sigma_i^2 + \tau w_i \cfrac{\nu}{\nu + 1} (\Delta_i^2 - \sigma_i^2)
    \nonumber \\
    &= \left ( 1 - \tau w_i \cfrac{\nu}{\nu + 1} \right ) \sigma_i^2 + \tau w_i \cfrac{\nu}{\nu + 1} \Delta_i^2
    \nonumber \\
    &= (1 - \tau_{\sigma_i}) \sigma_i^2 + \tau_{\sigma_i} \Delta_i^2
\end{align}
where $\tau_{\sigma_i} = \tau w_i \nu / (\nu + 1)$.

\begin{figure*}[tb]
    \centering
    \subfigure[Object of the target network]{
        \includegraphics[keepaspectratio=true,width=0.315\linewidth]{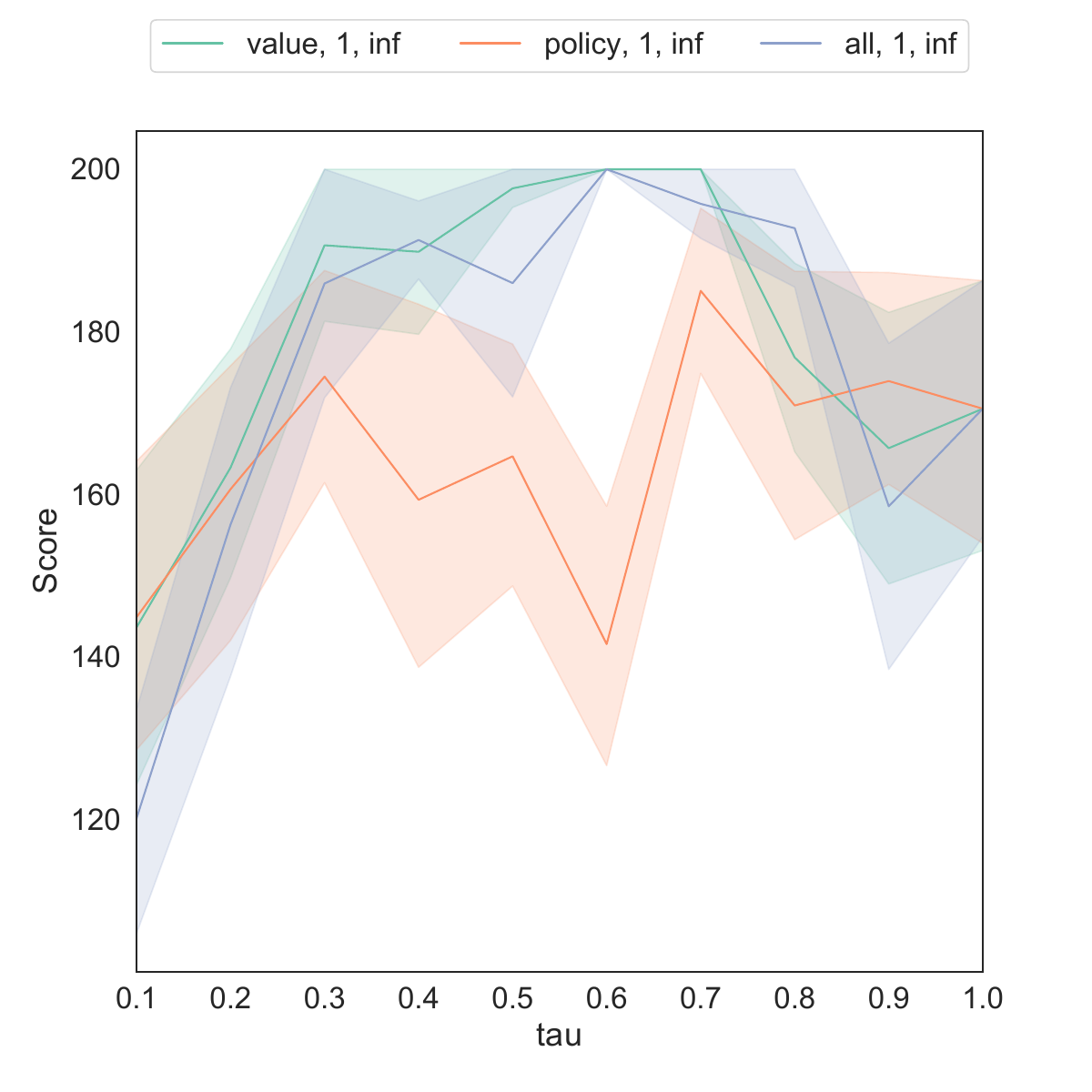}
    }
    \centering
    \subfigure[Update interval $n$]{
        \includegraphics[keepaspectratio=true,width=0.315\linewidth]{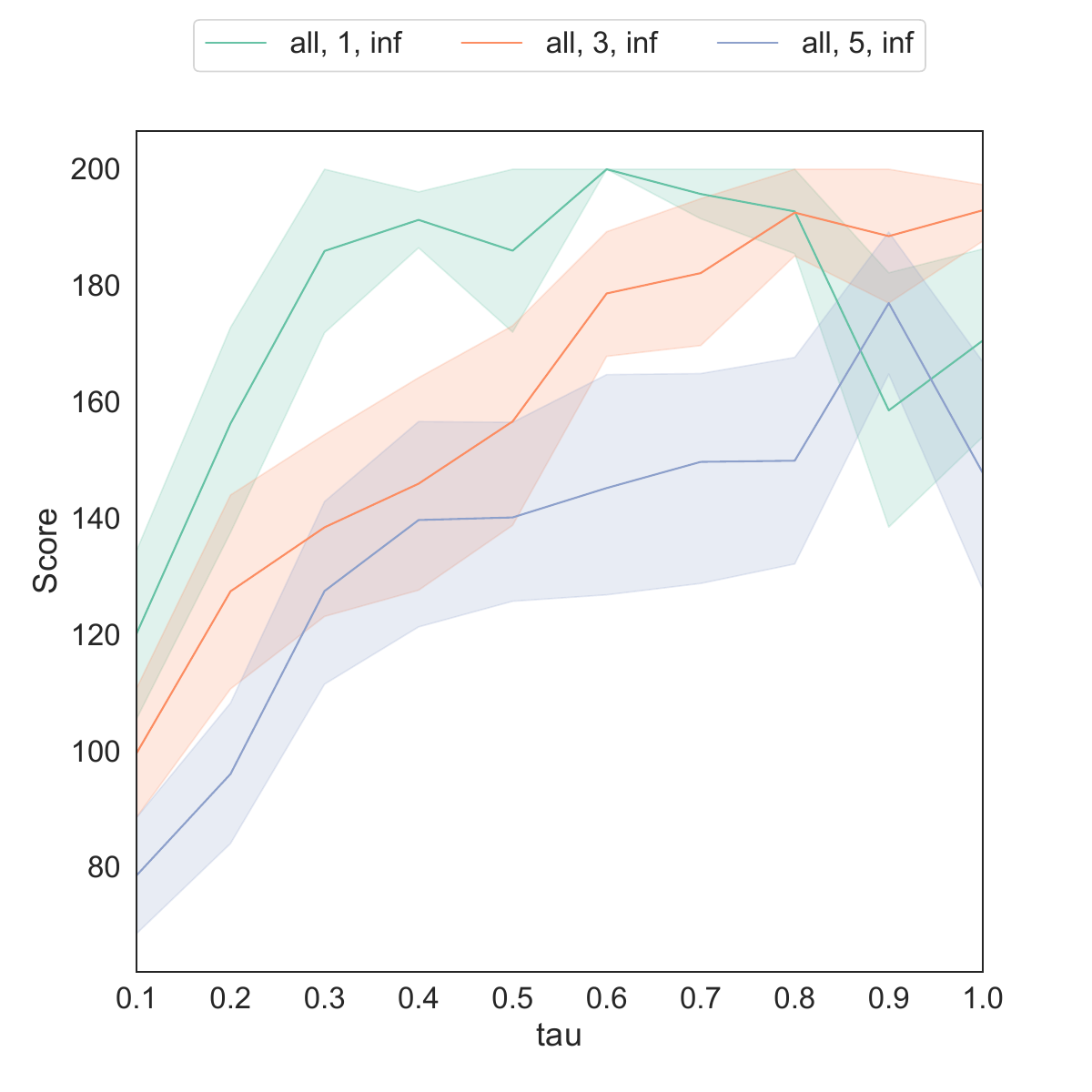}
    }
    \centering
    \subfigure[Degree of freedom $\nu$]{
        \includegraphics[keepaspectratio=true,width=0.315\linewidth]{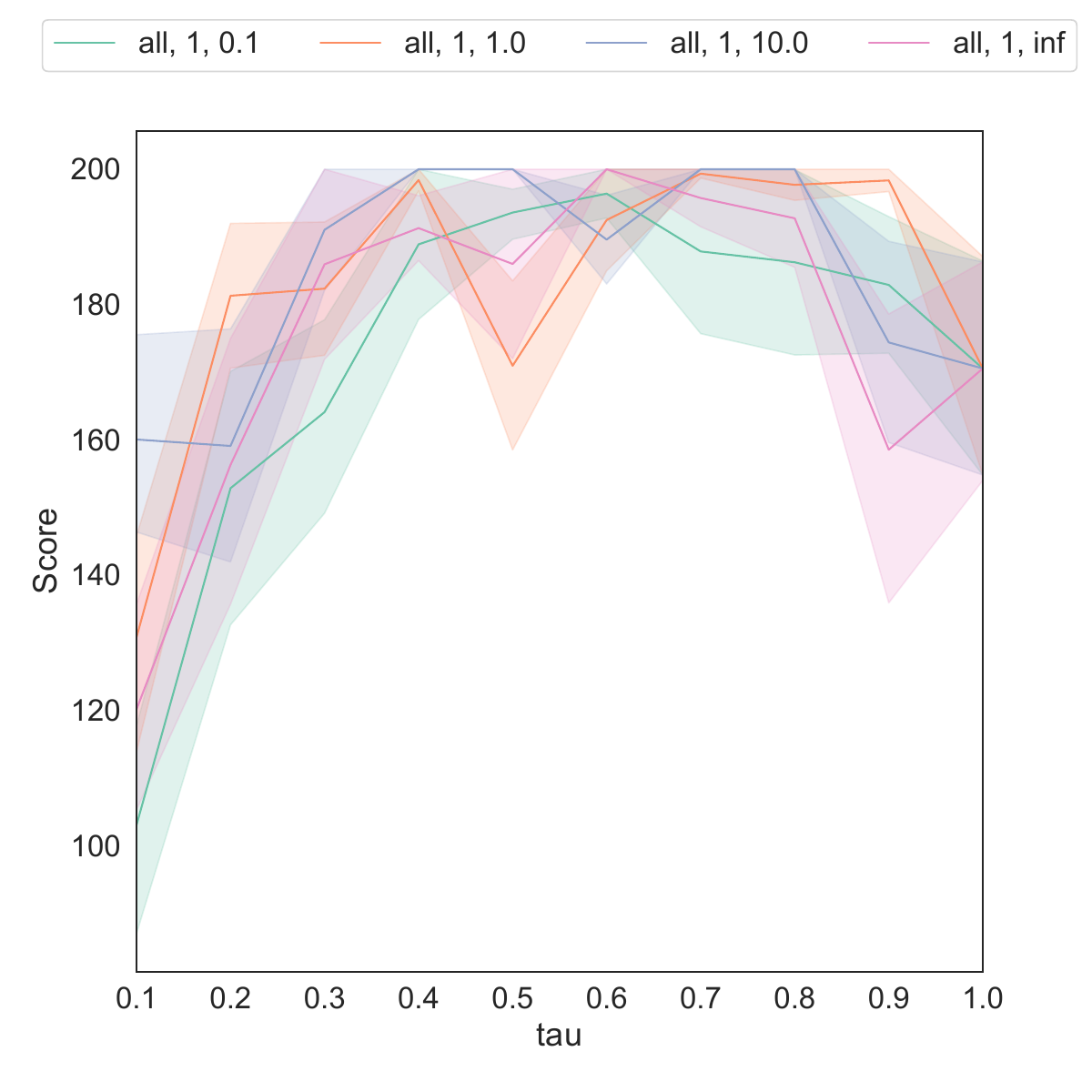}
    }
    \caption{Comparison results of learning CartPoleContinuousBulletEnv-v0:
        (a) reveals that the target network for value function is the most influential rather than the one for policy, but both of them would make a few improvements for increasing $\tau$;
        (b) shows that the deterioration of learning speed is not ignorable when $n > 1$ is used with the soft update, while the hard update with $\tau=1$ and $n=3$ has the similar performance to the soft update with $\tau \simeq 0.5$ and $n=1$;
        in (c), small $\nu$ tends to make the design of $\tau$ robust, although $\nu < 1$ seems to be too much robust to inhibit learning.
    }
    \label{fig:select_comp}
\end{figure*}

\section{Detailed results for selection of comparisons}
\label{app:select_comp}

A toy problem (i.e., CartPoleContinuousBulletEnv-v0) was solved 10 times for each condition with different random seeds.
Fig.~\ref{fig:select_comp} summarized the learning results.

As can be seen in Fig.~\ref{fig:select_comp}(a), the target network for the value function absolutely improves the learning performance in comparison with the one for the policy.
However, by applying it into both of them (as all), the learning performance seems to be improved especially when $\tau$ is large.

In Fig.~\ref{fig:select_comp}(b), the effects of the update interval $n$ are investigated when the target network is given for all the functions.
When $n > 1$, the learning performance was decreased as $\tau$ gets smaller.
That is, the combination of $n$ and $\tau$ causes a slowdown in the learning speed of the main network since the reference signals generated by the target network are updated too slowly.
Note, however, that the hard update with $\tau=1$ and $n=3$ achieves the same level of performance as the soft update with $\tau \simeq 0.5$ and $n=1$.
In the case of $\tau \simeq 0.5$, it takes about three steps for roughly 90\% copying the parameters to the target network from the main one.
This fact suggests that the three-step delay for the update of target network is desirable for the learning method used in the simulations.
Note that this value $\tau \simeq 0.5$ is quite different from the default value of the DRL libraries, which is developed as open source software~\citep{stooke2019rlpyt,fujita2019chainerrl}, but that difference is largely due to online learning in our method.

Lastly, Fig.~\ref{fig:select_comp}(c) indicates that, as $\nu$ decreases, the range of valid $\tau$ values is extended.
For $\nu < 1$, however, the learning performance was totally deteriorated probably because the update speed of target network is too slow.
In addition, we notice that the lack of significant improvement by the t-soft update is due to the simple task solved.

\section*{\refname}
\bibliographystyle{elsarticle-harv}
\bibliography{biblio}

\end{document}